\documentclass[twocolumn]{onethree}

\usepackage[toc,page,header]{appendix}
\usepackage{minitoc}
\usepackage{amsmath,amssymb}
\usepackage{bm}
\usepackage{wrapfig}
\usepackage{enumitem}

\usepackage{graphicx}
\usepackage{booktabs}
\usepackage{multirow}
\usepackage[table]{xcolor}
\definecolor{altrow}{HTML}{EEF9FB}

\usepackage{wrapfig}

\usepackage[table,dvipsnames]{xcolor} 
\usepackage[most]{tcolorbox}
\usepackage{colortbl}     
\definecolor{lightgray}{gray}{0.96} 
\usepackage{tabularx}
\usepackage{booktabs}
\usepackage{longtable}

\usepackage{arydshln}

\title{UnifiedAD: One-Shot Anomaly Detection across Modalities via Evidence Calibration}

\author[1,3]{Shangze Li*}
\author[2,3]{Chuancheng Shi*}
\author[2]{Simiao Xie}
\author[2]{Wenhua Wu}
\author[3]{Xingyu Zhu}
\author[3]{Fei Shen$^{\dagger}$}
\author[3]{Tat-Seng Chua}

\affiliation[1]{Nanjing University of Science and Technology}
\affiliation[2]{University of Sydney}
\affiliation[3]{National University of Singapore, NExT++ Research Centre}

\abstract{
Both single-modal and multi-modal anomaly detection aim to identify deviations from normal patterns, providing a common foundation for unified detection.
However, existing methods rely on modality-specific modeling and scoring, making it difficult to obtain comparable anomaly evidence with limited normal references.
To that end, we propose \textbf{UnifiedAD}, which unifies single-modal and multi-modal anomaly detection through anomaly evidence calibration, requiring only one normal reference per target category and no gradient updates.
Specifically, we first derive neuron-level discriminative priors from normal and anomalous samples in an independent auxiliary dataset and combine them with neuron response consistency under normality-preserving transformations of the target reference to select reliable neurons, termed anomaly neurons, and identify anomaly directions.
Next, we construct a low-rank normal subspace for each layer and modality from the selected neurons’ spatial responses to the reference and quantify query feature deviations from normal structure using projection residuals.
Finally, we separately calibrate projection residuals and anomaly-direction responses using the response distributions obtained from the normal reference and its transformations, converting them into comparable anomaly evidence for aggregation across layers and modalities.
Experiments on multiple anomaly detection benchmarks demonstrate that UnifiedAD achieves strong performance in both image-level anomaly detection and pixel-level anomaly localization.
\vspace{-0.1cm}
}

\checkdata[Resources]{
  \href{https://github.com/LucasShi1101/UnifiedAD}{\raisebox{-1ex}{\includegraphics[height=4.2ex]{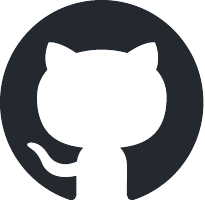}}~Code}\hspace{1em}
  \hspace{1em}\href{mailto:shangzeli@njust.edu.cn}{\nolinkurl{shangzeli@njust.edu.cn}}
  \hspace{1em}\href{mailto:cshi0459@uni.sydney.edu.au}{\nolinkurl{cshi0459@uni.sydney.edu.au}}
  \vspace{0.2cm}
}

\checkdata[Date]{
  November 24, 2026
}

\begin{document}
\maketitle

\begingroup
  \renewcommand{\thefootnote}{\fnsymbol{footnote}}
  \footnotetext[1]{Equal Contribution.}
\footnotetext[2]{Corresponding Author (\href{mailto:shenfei29@nus.edu.sg}{shenfei29@nus.edu.sg}).}
\endgroup

\section{Introduction}

Both single-modal and multi-modal anomaly detection~\cite{zhai2026foundad,li2021cutpaste,zhou2024anomalyclip,xu2026mrad,damm2025anomalydino} aim to identify deviations from normal patterns, providing a common foundation for unified detection.
Transferable representations from pretrained vision models~\cite{radford2021learning,oquab2023dinov2} provide a foundation for anomaly detection, enabling normal patterns of a target category to be characterized using few normal reference samples.
Therefore, developing a unified detection mechanism requiring only one normal reference can help reduce the data and adaptation requirements for new categories.

\begin{figure}[t]
\centering
\includegraphics[width=0.98\linewidth]{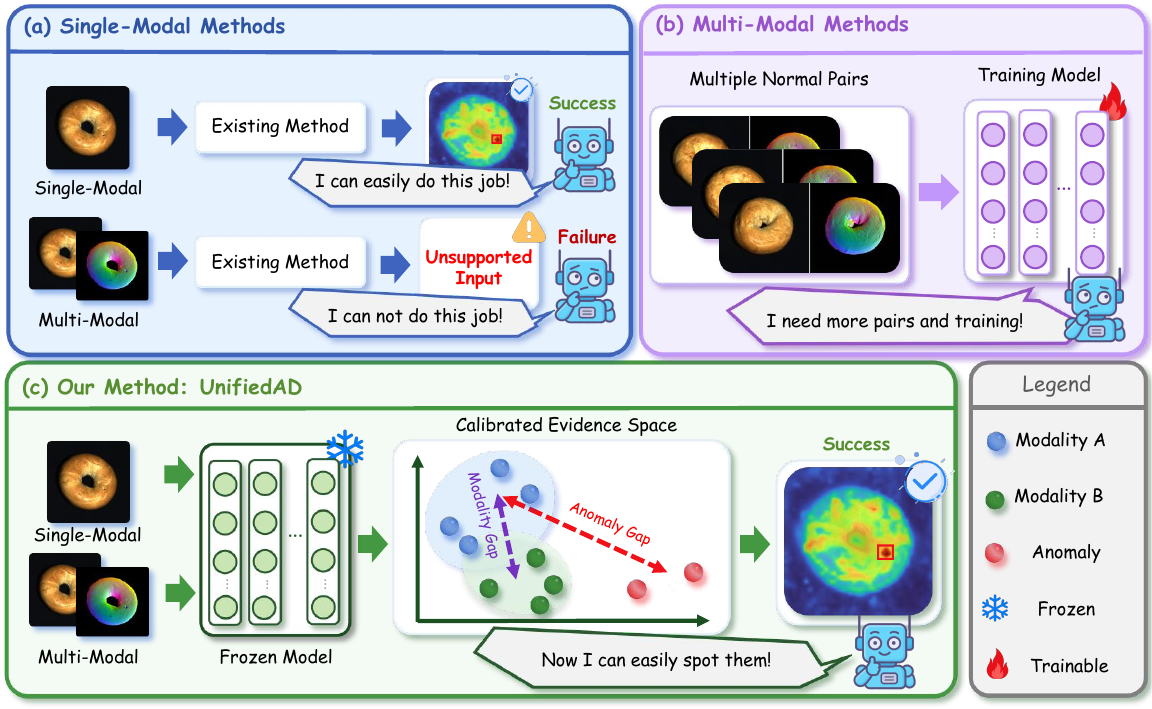}
\vspace{-0.2cm}
\caption{\textbf{Motivation for UnifiedAD.} (a) Single-modal methods typically lack direct multi-modal support. (b) Multi-modal methods often require multiple normal pairs and training. (c) UnifiedAD calibrates anomaly evidence for unified detection using one normal reference per target category.}
\label{fig:q1}
\vspace{-0.5cm}
\end{figure}

Existing methods~\cite{huang2022registration,zhu2024InCTRL,li2026gpflow} typically develop input-specific normality models and anomaly scoring mechanisms, limiting their ability to handle their inputs within a unified detection framework.
As shown in Figure~\ref{fig:q1}(a), single-modal methods~\cite{defard2021padim,deng2022anomaly,lendering2026subspacead,roth2022towards,yu2021fastflow,zhai2026foundad} typically construct feature memory banks, prototypes, or statistical distributions from normal samples and identify anomalies through feature matching or distance measures.
Multi-modal methods~\cite{cai2026cmds,costanzino2024cfm,guo2026cfr,lin2026cif,mao2025beyond,wang2023m3dm} exploit complementary information through feature fusion, cross-modal prediction, or reconstruction, as illustrated in Figure~\ref{fig:q1}(b).
With only one normal reference per target category, limited observations constrain normal structure estimation~\cite{li2025find,li2026fastref}, while differences in response scales and normal variability across modalities hinder the direct comparison and aggregation of raw deviations~\cite{li2025hgcf}.
Therefore, unified detection with a single normal reference must address two challenges: reliably modeling normal structure and constructing comparable anomaly evidence.

These challenges motivate us to view anomalies from a comparative perspective: whether a change is anomalous should be assessed relative to the corresponding normal reference.
Deviations of the same magnitude may be normal in a modality with high normal variability but anomalous in one with stable responses.
We therefore hypothesize that the rarity of a deviation under its corresponding normal response distribution can serve as a common scale for anomaly assessment across modalities.
This motivates us to retain the normal response structure of each modality while constructing comparable anomaly evidence relative to its normal variability, thereby unifying anomaly assessment across input settings.

To this end, we propose \textbf{UnifiedAD}, which unifies single-modal and multi-modal anomaly detection through anomaly evidence calibration, requiring only one normal reference per target category and no gradient updates.
Specifically, we first derive neuron-level discriminative priors from normal and anomalous samples in an independent auxiliary dataset and combine them with neuron response consistency under normality-preserving transformations of the target normal reference to select discriminative and stable neurons and identify anomaly directions.
We then construct a low-rank normal subspace for each layer and modality from the selected neurons' spatial responses to the reference and quantify query feature deviations from normal structure using projection residuals.
Finally, we use the response distributions obtained from the normal reference and its transformations to calibrate projection residuals and anomaly-direction responses separately, yielding comparable anomaly evidence that we aggregate across layers and modalities.
Experiments on multiple anomaly detection benchmarks demonstrate that UnifiedAD achieves strong performance in both image-level anomaly detection and pixel-level anomaly localization. Our main contributions are as follows:

\begin{itemize}
   \item We propose UnifiedAD, which leverages discriminative priors from auxiliary data to unify single-modal and multi-modal anomaly detection, requiring only one normal reference per target category and no gradient updates.

    \item We devise anomaly neuron selection, low-rank normal subspace modeling, and calibration based on normal variability to convert responses across modalities into comparable anomaly evidence for unified aggregation.

    \item We validate the effectiveness of UnifiedAD in image-level anomaly detection and pixel-level anomaly localization across multiple benchmarks and assess the contribution of each component through ablation studies.
\end{itemize}

\section{Related Work}

\noindent\textbf{Anomaly Detection with Pretrained Features.}
Alongside reconstruction-based approaches~\cite{deng2022anomaly,tien2023revisiting,zavrtanik2021draem}, methods based on pretrained features exploit transferable representations to model normal patterns and identify anomalies through feature deviations~\cite{zhai2026foundad,liu2023simplenet,batzner2024efficientad}.
In single-modal settings, representative methods~\cite{defard2021padim,lendering2026subspacead,roth2022towards} primarily adopt memory matching or statistical modeling.
PatchCore~\cite{roth2022towards} stores representative normal patch features and computes anomaly scores through nearest-neighbor retrieval, while PaDiM~\cite{defard2021padim} models multi-layer features at each spatial location with multivariate Gaussian distributions and uses Mahalanobis distance~\cite{mahalanobis2018generalized} for detection and localization.
For few-shot detection, SubspaceAD~\cite{lendering2026subspacead} constructs a normal subspace from frozen visual features using principal component analysis (PCA)~\cite{jolliffe2016principal}, enabling normal structure modeling in the one-shot setting.
In multi-modal settings, M3DM~\cite{wang2023m3dm} combines feature fusion and decision-level fusion through multiple memory banks, whereas CFM~\cite{costanzino2024cfm} learns cross-modal feature mappings and identifies anomalies through inconsistencies between observed and predicted features.
However, anomaly scoring in these methods~\cite{li2026gpflow,costanzino2026modulate} depends on their respective feature spaces and normality models, and raw deviations across layers and modalities are generally not directly comparable.

\noindent\textbf{Unified Anomaly Detection.}
Unified anomaly detection~\cite{you2022unified,he2024mambaad,guo2025dinomaly,luo2025INPformer,zhu2024InCTRL,gu2025univad} aims to handle different categories, domains, or task settings through shared models or inference mechanisms, although the scope of unification varies across studies.
UniAD~\cite{you2022unified}, MambaAD~\cite{he2024mambaad}, and Dinomaly~\cite{guo2025dinomaly} explore using a single model for multiple object categories, reducing the need for separate category-specific detectors.
INP-Former~\cite{luo2025INPformer} models intrinsic normal prototypes to support single-class, multi-class, and few-shot settings.
For cross-domain detection, InCTRL~\cite{zhu2024InCTRL} learns a transferable residual scoring model from labeled auxiliary data and performs inference using a few normal references from the target dataset, without further training.
UniVAD~\cite{gu2025univad} combines component segmentation, patch matching, and component relation modeling for cross-domain detection without target-domain training, including the single-normal-reference setting.
However, these methods do not directly unify single-modal and multi-modal inference using a single normal reference per target category.

\begin{figure*}[t]
\centering
\includegraphics[width=0.99\linewidth]{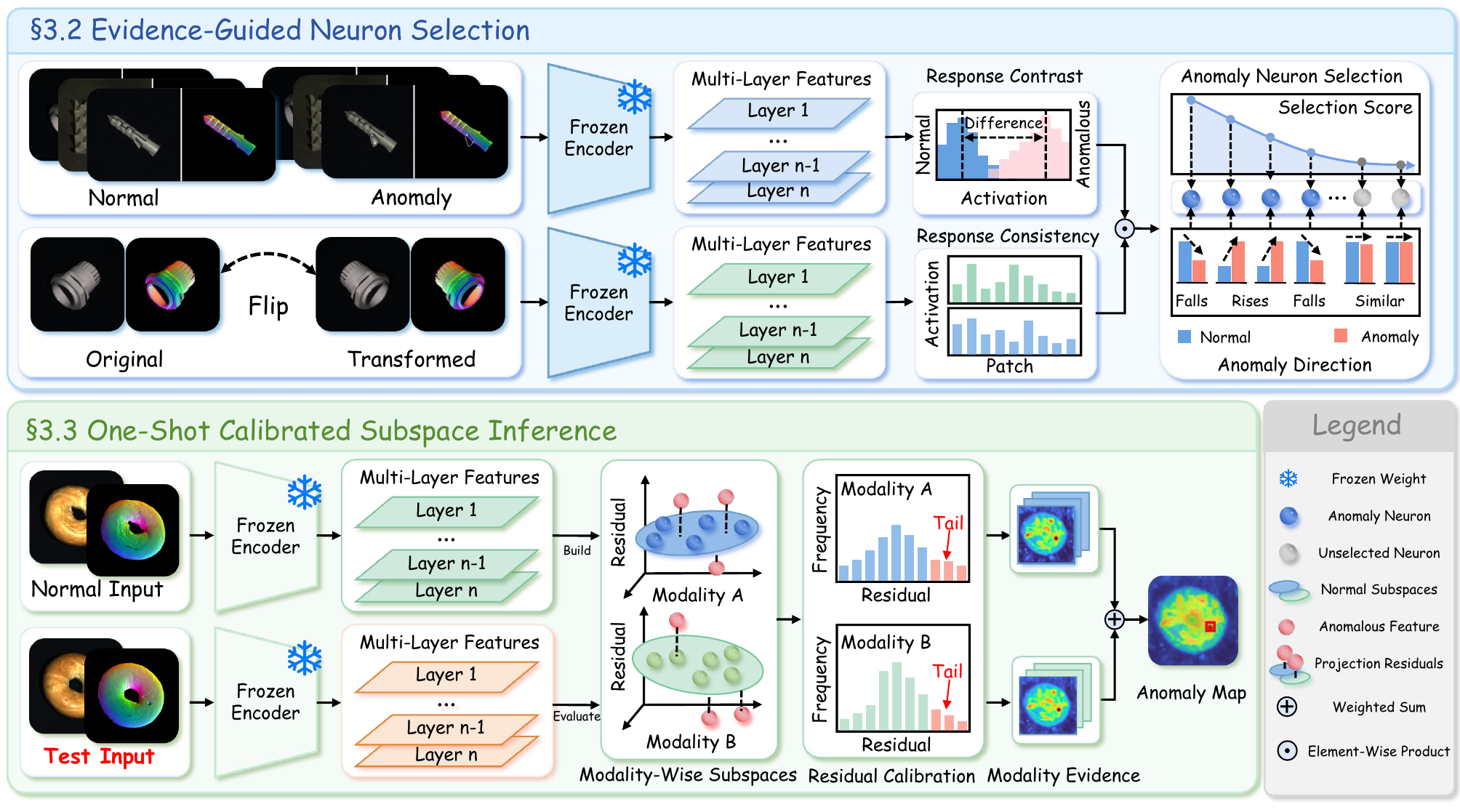}
\vspace{-0.3cm}
\caption{\textbf{Overview of UnifiedAD.} The framework consists of two stages: \textbf{(1) Evidence-Guided Neuron Selection}, which selects discriminative and stable neurons; and \textbf{(2) One-Shot Calibrated Subspace Inference}, which constructs normal subspaces and calibrates anomaly evidence for unified detection across modalities.}
\label{fig:pipeline}
\vspace{-0.5cm}
\end{figure*}

\section{Method}
\label{sec:method}

\subsection{Overall Framework}
\label{sec:overall_framework}

As shown in Figure~\ref{fig:pipeline}, we propose UnifiedAD, a unified one-shot anomaly detection framework for single-modal and multi-modal inputs. Given a target category $c$ and an active modality set $\mathcal{M}$, we denote the normal support and query samples by $\mathcal{S}_c=\{s_c^{(m)}\mid m\in\mathcal{M}\}$ and $x=\{x^{(m)}\mid m\in\mathcal{M}\}$, respectively. UnifiedAD extracts multi-layer features using frozen encoders and selects reliable neurons based on anomaly discriminability and transformation stability (Section~\ref{sec:egnc}). It then constructs low-rank normal subspaces from the support sample and calibrates query deviations into comparable anomaly evidence for unified aggregation (Section~\ref{sec:ocsi}).

\subsection{Evidence-Guided Neuron Selection}
\label{sec:egnc}

For each modality $m$, an input adapter $\mathcal{T}_m$ maps an arbitrary view $y^{(m)}$ into an encoder-compatible representation. A frozen encoder~\cite{oquab2023dinov2} then extracts multi-layer features:
\begin{equation}
    F_l^{(m)}(y)
    =\Phi_l^{(m)}\!\left(\mathcal{T}_m(y^{(m)})\right)
    \in\mathbb{R}^{P_l\times D_l},
    \qquad l\in\mathcal{L}_m,
    \label{eq:multimodal_features}
\end{equation}
where $\Phi_l^{(m)}$ denotes the output of the modality encoder at layer $l$, $\mathcal{L}_m$ is the set of selected layers, and $P_l$ and $D_l$ are the numbers of patches and feature neurons, respectively. 
For each modality $m$ and feature layer $l$, each neuron is evaluated using a unified reliability-weighted discriminability criterion. 
Let $e_{\mathrm{aux},l,j}^{(m)}$ denote the signed anomaly effect estimated from auxiliary normal and anomalous samples using activation shift, anomaly-localization consistency, and directional agreement across categories. We define its anomaly discriminability as $i_{l,j}^{(m)}=|e_{\mathrm{aux},l,j}^{(m)}|$.
To evaluate transformation stability, we apply a modality-valid normality-preserving transformation $\tau_m$ to the support observation and obtain $\widetilde{s}_c^{(m)}=\tau_m(s_c^{(m)})$. Let $F_{l,s}^{(m)}=F_l^{(m)}(s_c^{(m)})$ and $\widetilde F_{l,s}^{(m)}=F_l^{(m)}(\widetilde s_c^{(m)})$ denote the spatially aligned feature matrices. The transformation stability of neuron $j$ is defined as
\begin{equation}
    \eta_{l,j}^{(m)}
    =
    \frac{1}{
        \operatorname{MSE}_{p}\!\left(
            F_{l,s,p,j}^{(m)},
            \widetilde F_{l,s,p,j}^{(m)}
        \right)
        +\epsilon_l
    },
    \label{eq:transformation_stability}
\end{equation}
where $p\in\{1,\ldots,P_l\}$ indexes spatial patches and $\epsilon_l>0$ is a small constant added for numerical stability. Anomaly discriminability and transformation stability are combined into a unified neuron selection score:
\begin{equation}
    q_{l,j}^{(m)}
    =
    \mathcal{N}_{l}^{(m)}\!\left(
        i_{l,j}^{(m)} \eta_{l,j}^{(m)}
    \right),
    \label{eq:unified_certificate}
\end{equation}
where $\mathcal{N}_{l}^{(m)}$ denotes neuron-wise score normalization within layer $l$. The signed effect is sparsified and normalized to obtain the auxiliary anomaly direction
\begin{equation}
    w_l^{(m)}
    =
    \operatorname{SparseNormalize}\!\left(
        e_{\mathrm{aux},l}^{(m)}
        \odot \eta_l^{(m)}
    \right),
    \label{eq:auxiliary_certificate}
\end{equation}
where $\odot$ denotes element-wise multiplication. The selected anomaly neuron set is
\begin{equation}
    S_l^{(m)}
    =
    \operatorname{TopK}_{j}\!\left(
        q_{l,j}^{(m)}
    \right),
    \label{eq:certified_neurons}
\end{equation}
where $\operatorname{TopK}_j$ returns the indices of the $K$ highest-scoring neurons, which we refer to as anomaly neurons. This selection is performed independently for each active modality.

\begin{table*}[t]
  \centering
  \caption{
    \textbf{Single-Modal Anomaly Detection Results on MVTec-AD and VisA.}
    All values are percentages. Our results are reported as
    mean~$\pm$~standard deviation.
    Methods are ordered by MVTec-AD AUROC in ascending order
    within each evaluation level.
    The best result in each column is highlighted in
    \textbf{bold}, and our method is shaded in light green.
    N/A denotes metrics that are not applicable, while
    -- indicates unavailable results.
  }
  \vspace{-0.2cm}
  \label{tab:mvtec_visa_results}
  \resizebox{0.98\textwidth}{!}{%
    \begin{tabular}{clcccccccc}
      \toprule
      \multirow{2}{*}{\textbf{Level}}
        & \multirow{2}{*}{\textbf{Method}}
        & \multicolumn{4}{c}{\textbf{MVTec-AD~\cite{bergmann2019mvtec}}}
        & \multicolumn{4}{c}{\textbf{VisA~\cite{zou2022spot}}} \\
      \cmidrule(lr){3-6}\cmidrule(lr){7-10}
        &
        & AUROC $\uparrow$
        & AP $\uparrow$
        & $F_1$-max $\uparrow$
        & AUPRO $\uparrow$
        & AUROC $\uparrow$
        & AP $\uparrow$
        & $F_1$-max $\uparrow$
        & AUPRO $\uparrow$ \\
      \midrule
      \multirow{7}{*}{Image}
        & AdaCLIP~\cite{cao2024adaclip}
          & 89.2 & 96.4 & 90.6 & \textcolor{gray}{N/A}
          & 79.7 & 79.6 & 83.2 & \textcolor{gray}{N/A} \\
        & VisualAD~\cite{hou2026visualad}
          & 90.1 & 94.8 & 92.4 & \textcolor{gray}{N/A}
          & 83.1 & 86.8 & 81.4 & \textcolor{gray}{N/A} \\
        & WinCLIP~\cite{jeong2023winclip}
          & 93.1 & 96.5 & 93.7 & \textcolor{gray}{N/A}
          & 83.8 & 85.1 & 83.1 & \textcolor{gray}{N/A} \\
        & FoundAD~\cite{zhai2026foundad}
          & 96.1 & 97.9 & 96.6 & \textcolor{gray}{N/A}
          & 92.6 & 92.7 & 89.9 & \textcolor{gray}{N/A} \\
        & SubspaceAD~\cite{lendering2026subspacead}
          & 97.1 & 98.6 & 96.8 & \textcolor{gray}{N/A}
          & 93.4 & 92.6 & 89.7 & \textcolor{gray}{N/A} \\
        & DCP-SFR~\cite{jiang2026dcpsfr}
          & 97.3 & 98.8 & 96.7 & \textcolor{gray}{N/A}
          & 92.0 & 93.7 & 87.8 & \textcolor{gray}{N/A} \\

        & \cellcolor[HTML]{EAF7EB}\textbf{UnifiedAD (Ours)}
          & \cellcolor[HTML]{EAF7EB}\textbf{97.9$_{\pm 0.2}$}
          & \cellcolor[HTML]{EAF7EB}\textbf{99.0$_{\pm 0.1}$}
          & \cellcolor[HTML]{EAF7EB}\textbf{96.9$_{\pm 0.1}$}
          & \cellcolor[HTML]{EAF7EB}\textcolor{gray}{N/A}
          & \cellcolor[HTML]{EAF7EB}\textbf{96.3$_{\pm 0.1}$}
          & \cellcolor[HTML]{EAF7EB}\textbf{96.8$_{\pm 0.1}$}
          & \cellcolor[HTML]{EAF7EB}\textbf{92.9$_{\pm 0.1}$}
          & \cellcolor[HTML]{EAF7EB}\textcolor{gray}{N/A} \\
      \midrule
      \multirow{7}{*}{Pixel}
        & AdaCLIP~\cite{cao2024adaclip}
          & 88.7 & 41.0 & 43.4 & 47.6
          & 95.1 & 29.2 & 33.8 & 71.3 \\
        & VisualAD~\cite{hou2026visualad}
          & 91.3 & 45.4 & 47.4 & 88.6
          & 95.3 & 29.9 & 35.2 & 88.2 \\
        & WinCLIP~\cite{jeong2023winclip}
          & 95.2 & 18.2 & 55.9 & 87.1
          & 96.4 & 5.4 & 41.3 & 85.1 \\          
        & FoundAD~\cite{zhai2026foundad}
          & 96.8 & 56.8 & 57.8 & 92.8
          & \textbf{99.7} & 39.7 & 47.7 & \textbf{98.0} \\
        & DCP-SFR~\cite{jiang2026dcpsfr}
          & 96.9 & 61.2 & 60.7 & 92.7
          & 98.2 & 40.4 & 44.9 & 93.3 \\
        & SubspaceAD~\cite{lendering2026subspacead}
          & 97.5 & 59.6 & 60.9 & 93.7
          & 98.2 & 37.4 & 44.8 & 93.5 \\

        & \cellcolor[HTML]{EAF7EB}\textbf{UnifiedAD (Ours)}
          & \cellcolor[HTML]{EAF7EB}\textbf{97.8$_{\pm 0.1}$}
          & \cellcolor[HTML]{EAF7EB}\textbf{68.4$_{\pm 0.5}$}
          & \cellcolor[HTML]{EAF7EB}\textbf{64.9$_{\pm 0.6}$}
          & \cellcolor[HTML]{EAF7EB}\textbf{94.7$_{\pm 0.1}$}
          & \cellcolor[HTML]{EAF7EB}99.2$_{\pm 0.0}$
          & \cellcolor[HTML]{EAF7EB}\textbf{52.2$_{\pm 0.8}$}
          & \cellcolor[HTML]{EAF7EB}\textbf{54.2$_{\pm 0.5}$}
          & \cellcolor[HTML]{EAF7EB}95.1$_{\pm 0.2}$ \\
      \bottomrule
    \end{tabular}%
  }
\end{table*}

\subsection{One-Shot Calibrated Subspace Inference}
\label{sec:ocsi}
We gather the selected-neuron responses as
$X_l^{(m)}=F_{l,s}^{(m)}[:,S_l^{(m)}]$
and standardize them using the support statistics:
\begin{equation}
    Z_l^{(m)}=
    \frac{X_l^{(m)}-\mu_l^{(m)}}{\gamma_l^{(m)}},
\end{equation}
where $\mu_l^{(m)},\gamma_l^{(m)}\in\mathbb{R}^{K}$ denote the neuron-wise center and scale, respectively.
We perform an economy-size singular value decomposition (SVD)~\cite{golub1970SVD},
$Z_l^{(m)}=U_l^{(m)}\Sigma_l^{(m)}V_l^{(m)\top}$.
Given a variance-retention threshold $\rho_m\in(0,1]$,
we select the smallest rank $k_l^{(m)}$ whose cumulative
explained variance~\cite{jolliffe2016principal} reaches $\rho_m$. The resulting normal basis is
\begin{equation}
    B_l^{(m)}
    =\left(V_{l,1:k_l^{(m)}}^{(m)}\right)^\top,
\end{equation}
where $B_l^{(m)}\in\mathbb{R}^{k_l^{(m)}\times K}$.
Let $F_{l,x}^{(m)}=F_l^{(m)}(x^{(m)})$ denote the query feature matrix. We select $S_l^{(m)}$ and standardize the query features using the support statistics $\mu_l^{(m)}$ and $\gamma_l^{(m)}$, obtaining $z_{l,p}^{(m)}\in\mathbb{R}^{K}$ for each patch $p$. Its projection onto the normal basis is
\begin{equation}
    \widehat z_{l,p}^{(m)}
    =z_{l,p}^{(m)}B_l^{(m)\top}B_l^{(m)}.
\end{equation}
The corresponding normalized projection residual is
\begin{equation}
    r_{l,p}^{(m)}=
    \frac{1}{|S_l^{(m)}|}
    \left\|
        z_{l,p}^{(m)}-\widehat z_{l,p}^{(m)}
    \right\|_2^2.
    \label{eq:subspace_residual}
\end{equation}
We apply the same standardization and projection to the aligned calibration view to obtain the reference residuals
$\mathcal{R}_l^{(m)}=\{\widetilde r_{l,i}^{(m)}\}_{i=1}^{N_l^{(m)}}$.
The query residual is then transformed into empirical-tail evidence:
\begin{equation}
    E_{\mathrm{sub},l,p}^{(m)}
    =-\log\left(
        \frac{
            1+\sum_{i=1}^{N_l^{(m)}}
            \mathbb{I}\!\left(
                \widetilde r_{l,i}^{(m)}
                \geq r_{l,p}^{(m)}
            \right)
        }{N_l^{(m)}+1}
    \right),
    \label{eq:tail_evidence}
\end{equation}
where $\mathbb{I}(\cdot)$ is the indicator function.
The auxiliary anomaly-direction response for each query patch is calculated as
\begin{equation}
    d_{l,p}^{(m)}=
    \frac{1}{D_l}
    \sum_{j=1}^{D_l}
    \bar z_{l,p,j}^{(m)}w_{l,j}^{(m)},
    \label{eq:auxiliary_response}
\end{equation}
where $\bar z_{l,p}^{(m)}\in\mathbb{R}^{D_l}$ denotes the query feature standardized across all neurons. Applying Eq.~\eqref{eq:tail_evidence} to this response and its calibration reference yields the calibrated auxiliary-direction evidence $E_{\mathrm{aux},l,p}^{(m)}$.
After aligning the layer-wise evidence maps to a common patch grid, the evidence for modality $m$ is aggregated across layers:
\begin{equation}
    A_p^{(m)}=
    \frac{1}{|\mathcal{L}_m|}
    \sum_{l\in\mathcal{L}_m}
    \left[
        E_{\mathrm{sub},l,p}^{(m)}
        +\alpha_m E_{\mathrm{aux},l,p}^{(m)}
    \right],
    \label{eq:layer_aggregation}
\end{equation}
where $\alpha_m$ weights the auxiliary anomaly-direction evidence.
Let $\beta_m$ denote the fusion weight of modality $m$. 
The fused patch-level anomaly evidence is:
\begin{equation}
    A_p=
    \sum_{m\in\mathcal{M}}\beta_m A_p^{(m)}.
    \label{eq:unified_inference}
\end{equation}
We reshape and upsample $A_p$ to obtain the pixel-level anomaly map $A\in\mathbb{R}^{H\times W}$, and compute the image-level anomaly score as $a_{\mathrm{img}}=\max_{u,v}A_{u,v}$.

\begin{table*}[t]
  \centering
  \caption{\textbf{Multi-Modal Anomaly Detection Results on MVTec 3D-AD.}
  All values are percentages. Baselines are evaluated once; UnifiedAD is reported as
  mean~$\pm$~sample standard deviation over three support seeds. The best result in each column is in \textbf{bold}.}
  \vspace{-0.2cm}
  \label{tab:mvtec3d_multimodal_results}
  \resizebox{\textwidth}{!}{%
    \setlength{\tabcolsep}{12pt}%
    \begin{tabular}{lccccccc}
      \toprule
      \multirow{2}{*}{\textbf{Method}}
        & \multicolumn{3}{c}{\textbf{Image-Level}}
        & \multicolumn{4}{c}{\textbf{Pixel-Level}} \\
      \cmidrule(lr){2-4}\cmidrule(lr){5-8}
        & AUROC $\uparrow$ & AP $\uparrow$ & $F_1$-max $\uparrow$
        & AUROC $\uparrow$ & AP $\uparrow$ & $F_1$-max $\uparrow$ & AUPRO $\uparrow$ \\
      \midrule
      CFM~\cite{costanzino2024cfm}
        & 67.5 & 89.7 & 90.4 & 91.4 & 28.9 & 32.7 & 90.5 \\
      CIF~\cite{lin2026cif}
        & 72.0 & 88.0 & 89.4 & 86.1 & 12.5 & 17.6 & 82.9 \\
      M3DM~\cite{wang2023m3dm}
        & 73.9 & 91.2 & 90.7 & 90.2 & 28.0 & 32.1 & 90.2 \\
      CFR~\cite{guo2026cfr}
        & 74.0 & 89.8 & 90.0 & 92.3 & 31.1 & 35.6 & 90.4 \\
      CMDS-AD~\cite{cai2026cmds}
        & 79.6 & 88.9 & 90.5 & 94.2 & 30.8 & 32.6 & 88.4 \\
      \cellcolor[HTML]{EAF7EB}\textbf{UnifiedAD (Ours)}
        & \cellcolor[HTML]{EAF7EB}\textbf{92.8$_{\pm 0.3}$}
        & \cellcolor[HTML]{EAF7EB}\textbf{97.7$_{\pm 0.1}$}
        & \cellcolor[HTML]{EAF7EB}\textbf{93.9$_{\pm 0.1}$}
        & \cellcolor[HTML]{EAF7EB}\textbf{97.6$_{\pm 0.1}$}
        & \cellcolor[HTML]{EAF7EB}\textbf{62.5$_{\pm 0.9}$}
        & \cellcolor[HTML]{EAF7EB}\textbf{60.4$_{\pm 1.0}$}
        & \cellcolor[HTML]{EAF7EB}\textbf{97.5$_{\pm 0.1}$} \\
      \bottomrule
    \end{tabular}%
  }
  \vspace{-0.2cm}
\end{table*}

\section{Experiments and Analysis}

\noindent\textbf{Datasets.}
Following prior work~\cite{lendering2026subspacead, jiang2026dcpsfr, hou2026visualad}, we evaluate single-modal anomaly detection on MVTec-AD~\cite{bergmann2019mvtec} and VisA~\cite{zou2022spot} using one normal reference per category.
To keep auxiliary data separate from the evaluation benchmark, we adopt a cross-dataset protocol for neuron selection: discriminative priors are derived from VisA when evaluating on MVTec-AD and from MVTec-AD when evaluating on VisA.
For multi-modal industrial anomaly detection, we evaluate on all 10 categories of MVTec 3D-AD~\cite{bergmann2021mvtec3d}, using one spatially aligned normal RGB--XYZ pair per category.

\noindent\textbf{Metrics.}
Following the evaluation protocols~\cite{hou2026visualad, lendering2026subspacead}, we report the area under the receiver operating characteristic curve (AUROC), average precision (AP), and the maximum F1 score across thresholds ($F_1$-max) for image-level anomaly detection.
For pixel-level localization, we report the AUROC, AP, F1-max, and the area under the AUPRO.
On MVTec 3D-AD~\cite{bergmann2021mvtec3d}, we report AUROC, AP, and F1-max at the image level, and AUROC, AP, F1-max, and AUPRO at the pixel level. We repeat each one-shot UnifiedAD experiment three times with independently sampled normal references and report the mean~$\pm$~standard deviation.

\begin{figure}[t]
\centering
\includegraphics[width=\linewidth]{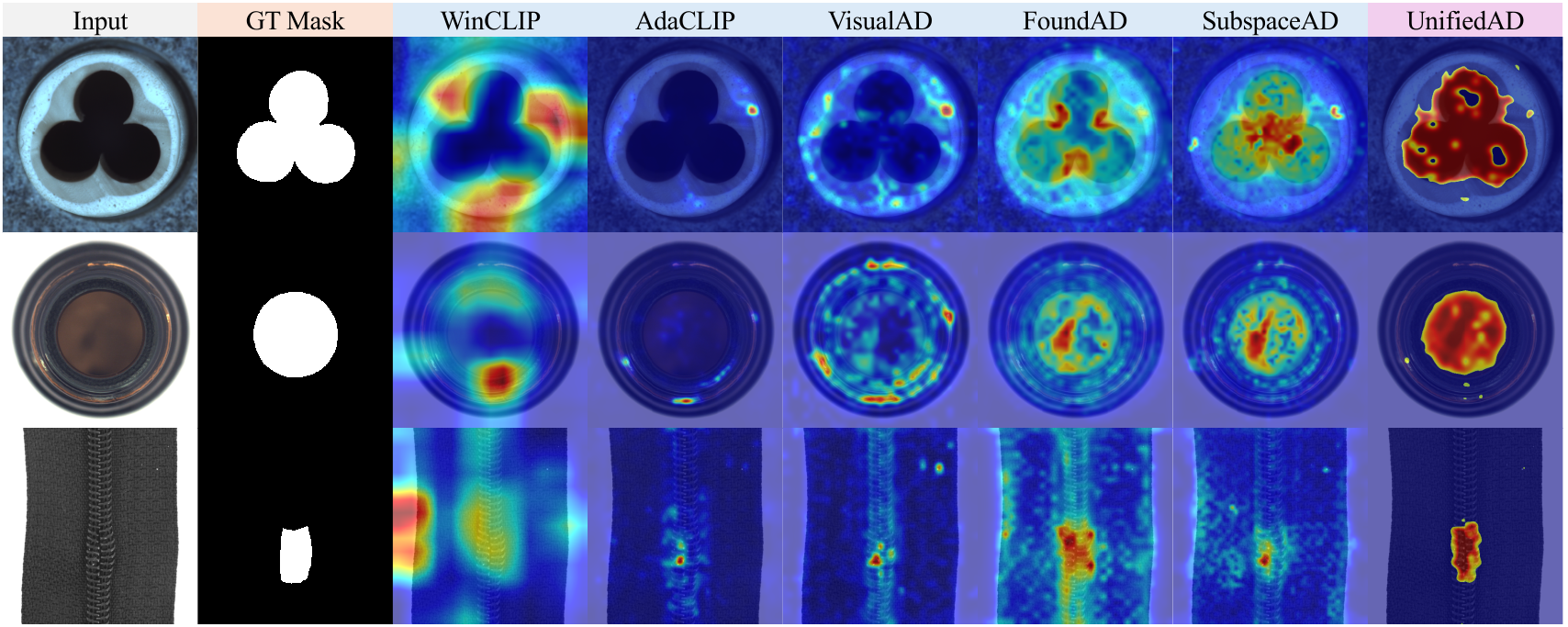}
\vspace{-0.2cm}
\caption{\textbf{Qualitative Comparison of Anomaly Localization Results on MVTec-AD.} UnifiedAD produces more accurate and concentrated anomaly responses with fewer false activations in normal regions compared with existing methods.}
\label{fig:single_dingxing}
\vspace{-0.3cm}
\end{figure}

\noindent\textbf{Hyperparameters.}
Following SubspaceAD~\cite{lendering2026subspacead}, we use a frozen DINOv2 ViT-g/14 (DINOv2-G) backbone~\cite{oquab2023dinov2} to encode both RGB images and geometry representations.
We select the top $K=128$ neurons per layer and set the retained-variance thresholds to $\rho_{\mathrm{RGB}}=0.90$ and $\rho_{\mathrm{G}}=0.95$.
The auxiliary-direction weights are $\alpha_{\mathrm{RGB}}=1$ and $\alpha_{\mathrm{G}}=0$, and RGB and geometry evidence receive equal weights during multi-modal fusion.
For organized XYZ inputs, we normalize depth using its 1st and 99th percentiles and concatenate it with normalized surface slopes along the $x$ and $y$ axes to form a three-channel geometry representation.
For BrainAD~\cite{bao2024brainad}, we use one normal slice as the reference and exclude all slices from the same patient from evaluation.
For MVTec 3D-AD~\cite{bergmann2021mvtec3d}, transferable and discriminative priors for neuron selection are derived from VisA.
Adaptation involves feature-statistic estimation and economy-size SVD without gradient updates.

\begin{figure}[t]
\centering
\includegraphics[width=\linewidth]{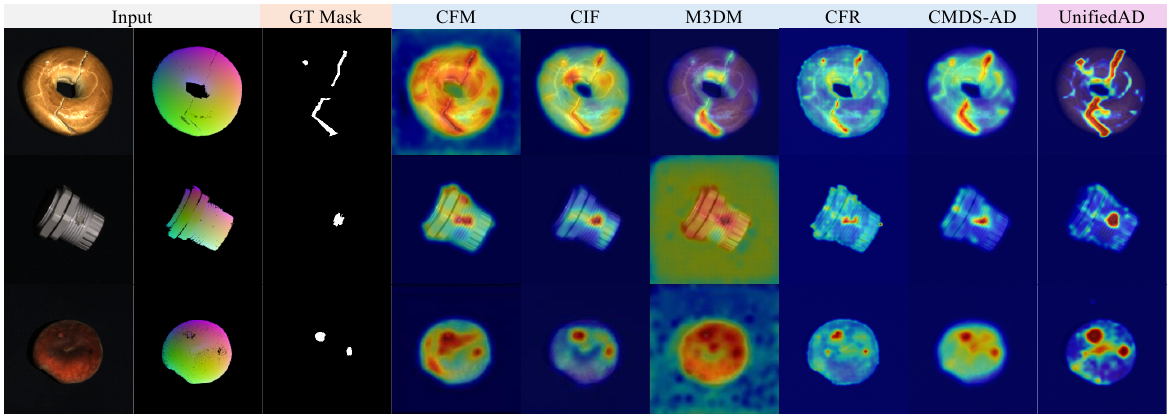}
\vspace{-0.2cm}
\caption{\textbf{Qualitative Comparison of Anomaly Localization Results on the MVTec 3D-AD.} UnifiedAD achieves more precise defect localization while effectively suppressing irrelevant responses across heterogeneous modalities.}
\label{fig:multi_dingxing}
\vspace{-0.3cm}
\end{figure}

\begin{table*}[t]
  \centering
  \caption{
    \textbf{Controlled Component Ablation.}
    Each variant changes one component of the full model while keeping all
    other settings fixed. Values are reported as mean~$\pm$~sample standard
    deviation. The best and second-best results
    are shown in \textbf{bold} and \underline{underlined}.
  }
  \label{tab:controlled_component_ablation}

  \footnotesize
  \setlength{\tabcolsep}{5pt}
  \renewcommand{\arraystretch}{1.12}

  \resizebox{\textwidth}{!}{%
    \begin{tabular}{@{}l*{9}{c}@{}}
      \toprule
      \multirow{2}{*}{\textbf{Variant}}
        & \multicolumn{3}{c}{\textbf{MVTec 3D-AD~\cite{bergmann2021mvtec3d}}}
        & \multicolumn{3}{c}{\textbf{MVTec-AD~\cite{bergmann2019mvtec}}}
        & \multicolumn{3}{c}{\textbf{VisA~\cite{zou2022spot}}} \\
      \cmidrule(lr){2-4}
      \cmidrule(lr){5-7}
      \cmidrule(l){8-10}
        & I-AUROC $\uparrow$ & P-AP $\uparrow$ & AUPRO $\uparrow$
        & I-AUROC $\uparrow$ & P-AP $\uparrow$ & AUPRO $\uparrow$
        & I-AUROC $\uparrow$ & P-AP $\uparrow$ & AUPRO $\uparrow$ \\
      \midrule

      B$_0$
        & 67.1$_{\pm0.2}$ & 28.7$_{\pm0.0}$ & 82.3$_{\pm0.1}$
        & 83.3$_{\pm0.4}$ & 43.2$_{\pm0.2}$ & 84.9$_{\pm0.1}$
        & 74.2$_{\pm0.5}$ & 30.3$_{\pm0.0}$ & 85.5$_{\pm0.1}$ \\
      \addlinespace[2pt]

      B$_1$
        & 78.2$_{\pm2.8}$ & 40.2$_{\pm1.0}$ & 91.3$_{\pm0.2}$
        & 84.3$_{\pm0.9}$ & 51.7$_{\pm0.5}$ & 84.1$_{\pm1.2}$
        & 82.6$_{\pm0.7}$ & 25.2$_{\pm0.4}$ & 82.1$_{\pm0.6}$ \\
      \addlinespace[2pt]

      B$_2$
        & 80.1$_{\pm0.7}$ & 44.8$_{\pm0.6}$ & 94.1$_{\pm0.4}$
        & \textbf{98.1$_{\pm0.2}$} & 65.3$_{\pm0.7}$ & 93.8$_{\pm0.1}$
        & \textbf{97.6$_{\pm1.8}$} & 50.7$_{\pm0.5}$ & \underline{94.8$_{\pm0.3}$} \\
      \addlinespace[2pt]
      B$_3$
        & \underline{84.3$_{\pm2.1}$} & \underline{51.8$_{\pm2.5}$} & \underline{96.4$_{\pm0.3}$}
        & 95.5$_{\pm0.4}$ & \textbf{71.9$_{\pm1.1}$} & \underline{94.5$_{\pm0.4}$}
        & 93.8$_{\pm3.3}$ & \textbf{56.9$_{\pm0.4}$} & 94.2$_{\pm0.6}$ \\

      \midrule
      \rowcolor[HTML]{EAF7EB}
      \textbf{Ours}
        & \textbf{92.8$_{\pm0.3}$} & \textbf{62.5$_{\pm0.9}$} & \textbf{97.5$_{\pm0.1}$}
        & \underline{97.9$_{\pm0.2}$} & \underline{68.4$_{\pm0.5}$} & \textbf{94.7$_{\pm0.1}$}
        & \underline{96.3$_{\pm0.1}$} & \underline{52.2$_{\pm0.8}$} & \textbf{95.1$_{\pm0.2}$} \\
      \bottomrule
    \end{tabular}%
  }
  \vspace{-0.2cm}
\end{table*}

\begin{table*}[t]
  \centering
  \caption{
    \textbf{Neuron Specificity Analysis.} Comparison of Original, Random, and evidence-guided neuron selection under single-modal and multi-modal settings. All results are reported as mean ± standard deviation, with the best results highlighted in \textbf{bold}.
  }
  \vspace{-0.2cm}
  \label{tab:single_multi_modal_results}

  \resizebox{0.98\textwidth}{!}{%
    \begin{tabular}{clccccccc}
      \toprule
      \multirow{2}{*}{\textbf{Setting}}
        & \multirow{2}{*}{\textbf{Method}}
        & \multicolumn{3}{c}{\textbf{Image-Level}}
        & \multicolumn{4}{c}{\textbf{Pixel-Level}} \\
      \cmidrule(lr){3-5}
      \cmidrule(lr){6-9}
        &
        & I-AUROC $\uparrow$
        & I-AP $\uparrow$
        & I-$F_1$-max $\uparrow$
        & P-AUROC $\uparrow$
        & P-AP $\uparrow$
        & P-$F_1$-max $\uparrow$
        & AUPRO $\uparrow$ \\
      \midrule

      \multirow{3}{*}{\shortstack{Single-Modal}}
        & Original~\cite{oquab2023dinov2}
          & 81.8 {$_{\pm0.6}$}
          & 90.2 {$_{\pm0.3}$}
          & 91.0 {$_{\pm0.2}$}
          & 91.6 {$_{\pm0.2}$}
          & 37.4 {$_{\pm1.2}$}
          & 42.6 {$_{\pm1.0}$}
          & 80.1 {$_{\pm0.7}$} \\
        & Random
          & 80.9 {$_{\pm1.2}$}
          & 90.1 {$_{\pm0.6}$}
          & 89.9 {$_{\pm0.5}$}
          & 89.4 {$_{\pm0.4}$}
          & 29.5 {$_{\pm2.0}$}
          & 35.8 {$_{\pm1.6}$}
          & 75.5 {$_{\pm1.5}$} \\
        & \cellcolor[HTML]{EAF7EB}\textbf{UnifiedAD (Ours)}
          & \cellcolor[HTML]{EAF7EB}\textbf{97.9 {$_{\pm0.2}$}}
          & \cellcolor[HTML]{EAF7EB}\textbf{99.0 {$_{\pm0.1}$}}
          & \cellcolor[HTML]{EAF7EB}\textbf{96.9 {$_{\pm0.1}$}}
          & \cellcolor[HTML]{EAF7EB}\textbf{97.8 {$_{\pm0.1}$}}
          & \cellcolor[HTML]{EAF7EB}\textbf{68.4 {$_{\pm0.5}$}}
          & \cellcolor[HTML]{EAF7EB}\textbf{64.9 {$_{\pm0.6}$}}
          & \cellcolor[HTML]{EAF7EB}\textbf{94.7 {$_{\pm0.1}$}} \\

      \midrule
      \multirow{3}{*}{\shortstack{Multi-Modal}}
        & Original~\cite{oquab2023dinov2}
          & 60.6 {$_{\pm1.4}$}
          & 85.1 {$_{\pm0.9}$}
          & 89.1 {$_{\pm0.3}$}
          & 94.8 {$_{\pm0.1}$}
          & 12.7 {$_{\pm1.0}$}
          & 16.8 {$_{\pm0.7}$}
          & 82.9 {$_{\pm0.4}$} \\
        & Random
          & 61.0 {$_{\pm2.4}$}
          & 85.4 {$_{\pm1.1}$}
          & 89.2 {$_{\pm0.2}$}
          & 94.7 {$_{\pm0.2}$}
          & 13.2 {$_{\pm0.9}$}
          & 17.1 {$_{\pm0.8}$}
          & 82.8 {$_{\pm0.6}$} \\
        & \cellcolor[HTML]{EAF7EB}\textbf{UnifiedAD (Ours)}
          & \cellcolor[HTML]{EAF7EB}\textbf{92.8 {$_{\pm0.3}$}}
          & \cellcolor[HTML]{EAF7EB}\textbf{97.7 {$_{\pm0.1}$}}
          & \cellcolor[HTML]{EAF7EB}\textbf{93.9 {$_{\pm0.1}$}}
          & \cellcolor[HTML]{EAF7EB}\textbf{97.6 {$_{\pm0.1}$}}
          & \cellcolor[HTML]{EAF7EB}\textbf{62.5 {$_{\pm0.9}$}}
          & \cellcolor[HTML]{EAF7EB}\textbf{60.4 {$_{\pm1.0}$}}
          & \cellcolor[HTML]{EAF7EB}\textbf{97.5 {$_{\pm0.1}$}} \\

      \bottomrule
    \end{tabular}%
  }
  \vspace{-0.3cm}
\end{table*}

\subsection{Quantitative Comparison with SOTA Methods}

\noindent\textbf{Single-Modal.} 
To evaluate UnifiedAD in single-modal anomaly detection, we conduct experiments on MVTec-AD~\cite{bergmann2019mvtec} and VisA~\cite{zou2022spot} benchmarks under the one-shot setting. As shown in Table~\ref{tab:mvtec_visa_results}, UnifiedAD achieves the best performance across all image-level metrics and most pixel-level metrics on both benchmarks. Specifically, UnifiedAD achieves pixel-level AP scores of 68.4\% and 52.2\% on MVTec-AD and VisA, outperforming the strongest baselines by 7.2 and 11.8 percentage points, respectively. These results demonstrate the effectiveness of UnifiedAD in modeling normal feature structures and identifying anomalous deviations using only a single normal reference sample.

\noindent\textbf{Multi-Modal.}
To evaluate the effectiveness of UnifiedAD in multi-modal anomaly detection, we conduct experiments on the MVTec 3D-AD~\cite{bergmann2021mvtec3d} benchmark under the one-shot setting. As shown in Table~\ref{tab:mvtec3d_multimodal_results}, UnifiedAD consistently outperforms existing baseline methods across all image-level and pixel-level evaluation metrics. Specifically, UnifiedAD achieves image-level AUROC and AP scores of 92.8\% and 97.7\%, outperforming the strongest baselines by 13.2 and 6.5 percentage points, respectively, while achieving a pixel-level AP of 62.5\%, exceeding the strongest baseline by 31.4 percentage points. These results demonstrate that UnifiedAD effectively integrates calibrated anomaly evidence from heterogeneous modalities, enabling accurate anomaly detection and localization using one normal reference pair per target category.

\subsection{Qualitative Comparison with SOTA Methods}

\noindent\textbf{Single-Modal.} 
To qualitatively evaluate the anomaly localization capability of UnifiedAD in single-modal settings, we visualize the anomaly maps produced by different methods. As shown in Figure~\ref{fig:single_dingxing}, existing methods often exhibit scattered anomalous responses in normal regions or fail to accurately delineate defect boundaries. In contrast, UnifiedAD produces more concentrated anomaly responses that closely align with the ground-truth masks while effectively suppressing background interference. These results demonstrate the effectiveness of UnifiedAD in distinguishing anomalous deviations from normal variations, enabling more precise anomaly localization.

\noindent\textbf{Multi-Modal.} 
To qualitatively evaluate the anomaly localization capability of UnifiedAD in multi-modal settings, we visualize the anomaly maps produced by different methods on the MVTec 3D-AD~\cite{bergmann2021mvtec3d} benchmark. As shown in Figure~\ref{fig:multi_dingxing}, existing methods often produce diffuse anomaly responses or exhibit false activations in normal regions, making it difficult to precisely localize defects. In contrast, UnifiedAD generates more concentrated anomaly responses that better align with the ground-truth masks while suppressing irrelevant background activations. These results demonstrate the effectiveness of UnifiedAD in integrating anomaly evidence from heterogeneous modalities for anomaly localization.

\subsection{Ablation Study}

\noindent\textbf{Component Ablation.}
Table~\ref{tab:controlled_component_ablation} evaluates four variants, each modifying one component of the full model while keeping all other settings fixed.
B$_0$ replaces low-rank normal subspace modeling with prototype matching;
B$_1$ uses a single calibrated feature layer instead of multi-layer aggregation;
B$_2$ removes empirical-tail calibration;
and B$_3$ excludes auxiliary anomaly-direction evidence.
Compared with B$_0$, the full model improves image-level AUROC from 67.1\% to 92.8\% on MVTec 3D-AD, from 83.3\% to 97.9\% on MVTec-AD, and from 74.2\% to 96.3\% on VisA, supporting the effectiveness of low-rank normal modeling.
Compared with B$_1$, multi-layer evidence aggregation raises pixel-level AP from 40.2\% to 62.5\%, from 51.7\% to 68.4\%, and from 25.2\% to 52.2\% on these datasets, respectively, demonstrating the benefit of combining evidence across layers.
Compared with B$_2$, empirical-tail calibration improves pixel-level AP and AUPRO across all three datasets.
The gains are particularly pronounced on MVTec 3D-AD, where image-level AUROC increases from 80.1\% to 92.8\% and pixel-level AP rises from 44.8\% to 62.5\%.
However, image-level AUROC decreases on MVTec-AD and VisA, indicating a trade-off between detection and localization metrics.
Compared with B$_3$, auxiliary direction evidence improves image-level AUROC by 8.5, 2.4, and 2.5 percentage points on MVTec 3D-AD, MVTec-AD, and VisA, respectively.
It also increases pixel-level AP from 51.8\% to 62.5\% on MVTec 3D-AD, while reducing pixel-level AP on the two single-modal benchmarks.
These results show that direction evidence complements calibrated subspace evidence for image-level detection, with localization trade-offs in single-modal settings.

\begin{table}[t]
    \centering
    \small
    \caption{
        \textbf{Direct Subspace Fusion Analysis.}
    Comparison of direct subspace fusion and Full UnifiedAD on MVTec 3D-AD. Results are reported as mean~$\pm$~standard deviation over three support seeds. All values are percentages.
    }
    \label{tab:cross_modal_calibration}
    \vspace{-0.2cm}
    \setlength{\tabcolsep}{5pt}
    \renewcommand{\arraystretch}{1.10}
    \resizebox{\columnwidth}{!}{%
    \begin{tabular}{lccc}
        \toprule
        \textbf{Method}
        & I-AUROC $\uparrow$
        & P-AP $\uparrow$
        & AUPRO $\uparrow$ \\
        \midrule
        
        Direct Subspace Fusion
        & 76.5$_{\pm1.0}$
        & 39.7$_{\pm0.5}$
        & 92.4$_{\pm0.4}$ \\
        
        \cellcolor[HTML]{EAF7EB}\textbf{UnifiedAD (Ours)}
        & \cellcolor[HTML]{EAF7EB}\textbf{92.8$_{\pm0.3}$}
        & \cellcolor[HTML]{EAF7EB}\textbf{62.5$_{\pm0.9}$}
        & \cellcolor[HTML]{EAF7EB}\textbf{97.5$_{\pm0.1}$} \\
        \bottomrule
    \end{tabular}%
    }
    \vspace{-0.3cm}
\end{table}

\noindent\textbf{Neuron Specificity Analysis.}
To investigate whether anomaly detection relies on specific neurons rather than arbitrary feature responses, we conduct a controlled ablation study comparing the proposed anomaly neurons with original and random under both single-modal and multi-modal settings. As shown in Table~\ref{tab:single_multi_modal_results}, UnifiedAD consistently outperforms both alternatives across all image-level and pixel-level metrics, indicating that random neuron selection is insufficient for reliable anomaly detection. Specifically, compared with Original, UnifiedAD improves image-level AUROC from 81.8\% to 97.9\% and pixel-level AP from 37.4\% to 68.4\% in the single-modal setting, while achieving improvements from 60.6\% to 92.8\% and from 12.7\% to 62.5\%, respectively, in the multi-modal setting. These results demonstrate that not all neurons contribute equally to anomaly detection, and selecting neurons with high anomaly discriminability and transformation stability is essential for accurate anomaly detection.

\noindent\textbf{Fusion Analysis. }
To investigate multi-modal anomaly detection, we compare Full UnifiedAD with Direct Subspace~\cite{lendering2026subspacead} Fusion, which directly aggregates independently computed subspace residuals from the RGB and XYZ modalities through weighted summation without prior calibration, on MVTec 3D-AD. As shown in Table~\ref{tab:cross_modal_calibration}, direct fusion yields substantially lower performance in both image-level anomaly detection and pixel-level anomaly localization. Specifically, Full UnifiedAD improves I-AUROC from 76.5\% to 92.8\%, P-AP from 39.7\% to 62.5\%, and AUPRO from 92.4\% to 97.5\%. These results indicate that directly fusing modality-specific subspace residuals is insufficient for reliable multi-modal detection, supporting the need for evidence calibration before fusion.

\begin{figure}[t]
\centering
\includegraphics[width=0.96\linewidth]{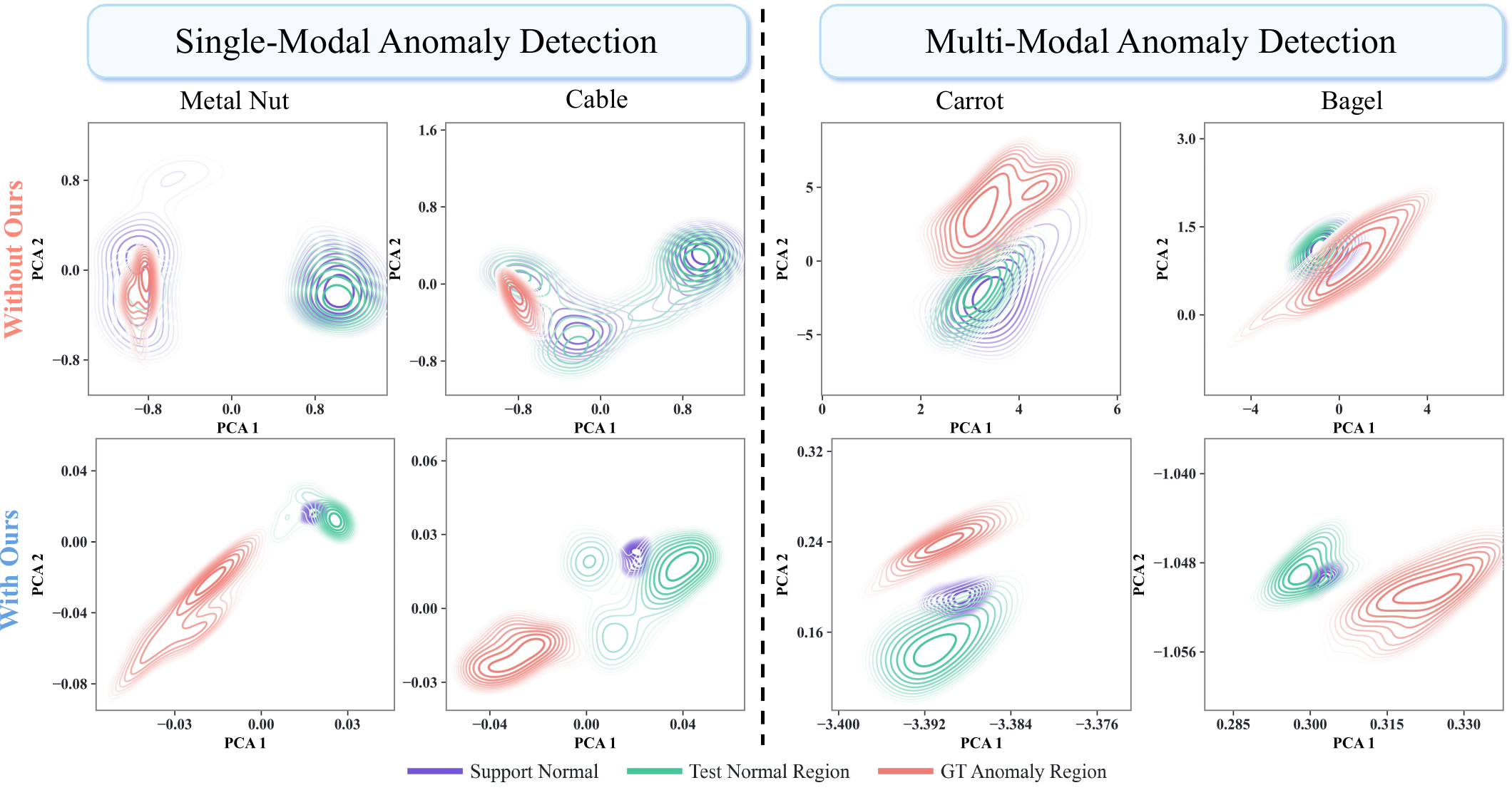}
\vspace{-0.3cm}
\caption{\textbf{Feature Space Visualization.} PCA visualization of normal and anomalous feature distributions before and after UnifiedAD under single-modal and multi-modal settings.}
\label{fig:tsne}
\vspace{-0.5cm}
\end{figure}

\noindent\textbf{Visualization.}
To investigate whether UnifiedAD can effectively distinguish anomalous deviations from normal feature structures, we visualize the feature distributions before and after our method using PCA under both single-modal and multi-modal settings. As shown in Figure~\ref{fig:tsne}, normal and anomalous features exhibit substantial overlap in the feature space before applying UnifiedAD, making it difficult to distinguish anomalous responses from normal variations. In contrast, after applying UnifiedAD, anomalous features become more clearly separated from normal features, while normal reference and test features remain relatively consistent. These results demonstrate that UnifiedAD enhances the discriminability between normal and anomalous responses, enabling more reliable anomaly detection across different input settings.

\begin{figure*}[t]
\centering
\includegraphics[width=0.96\linewidth]{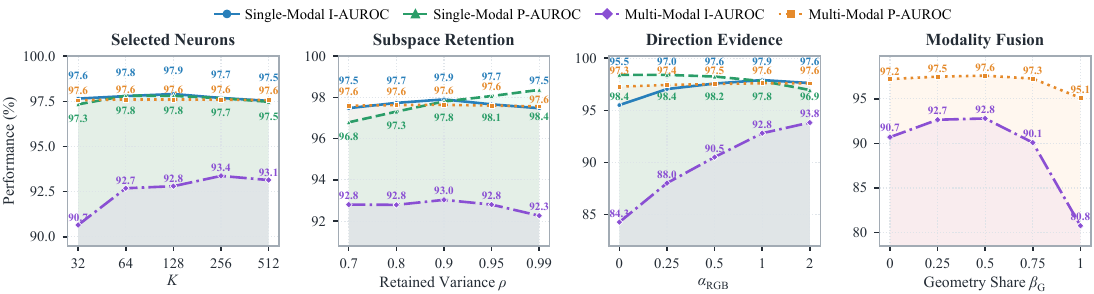}
\vspace{-0.3cm}
\caption{\textbf{Hyperparameter Sensitivity Analysis.} Effects of top-\(K\) neuron selection, subspace retention threshold \(\rho\), direction evidence weight \(\alpha_{\mathrm{RGB}}\), and geometry fusion weight \(\beta_G\) on image-level and pixel-level AUROC (\%) under single-modal and multi-modal settings.}
\label{fig:para}
\vspace{-0.2cm}
\end{figure*}

\begin{figure}[t]
\centering
\includegraphics[width=0.98\linewidth]{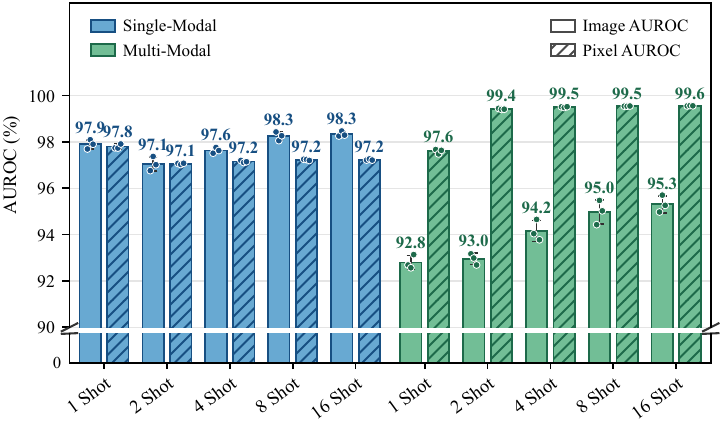}
\vspace{-0.3cm}
\caption{\textbf{Shot Number Ablation.} Image-level and pixel-level AUROC (\%) under single-modal and multi-modal settings with 1 to 16 normal reference samples.}
\label{fig:shot}
\vspace{-0.5cm}
\end{figure}

\subsection{Further Analysis}

\noindent\textbf{Impact of Neuron Selection Size.}
To evaluate the effect of the number of selected neurons, we conduct a sensitivity analysis by varying top-$K$ from 32 to 512. As shown in Figure~\ref{fig:para}, UnifiedAD maintains stable performance across different values of top-$K$. Specifically, the single-modal image-level AUROC reaches 97.9\% at top-$K$=128, while the multi-modal image-level AUROC improves from 90.7\% to 93.4\% at top-$K$=256. These results indicate that selecting a moderate number of anomaly neurons is sufficient to capture discriminative feature responses, and we therefore adopt top-$K$=128 as the default setting.

\noindent\textbf{Effect of Subspace Retention.} 
To examine the effect of variance retention on normal subspace modeling, we conduct a sensitivity analysis by varying \(\rho\) from 0.7 to 0.99. As shown in Figure~\ref{fig:para}, UnifiedAD maintains stable performance across different retention thresholds. Specifically, the single-modal pixel-level AUROC increases from 96.8\% to 98.4\%, while the multi-modal image-level AUROC remains around 92.8\%. These results indicate that the normal subspace can effectively characterize normal feature structures across different variance-retention thresholds.

\noindent\textbf{Influence of Direction Evidence Weight.} 
To evaluate the contribution of auxiliary anomaly-direction evidence, we conduct a sensitivity analysis by varying \(\alpha_{\mathrm{RGB}}\) from 0 to 2. As shown in Figure~\ref{fig:para}, increasing the weight of direction evidence improves image-level anomaly detection while maintaining stable pixel-level localization performance. Specifically, the multi-modal image-level AUROC increases from 84.3\% to 93.8\%, while the single-modal image-level AUROC reaches 97.9\% at \(\alpha_{\mathrm{RGB}}=1\). These results demonstrate that auxiliary anomaly-direction evidence effectively complements calibrated subspace evidence for more accurate anomaly detection.

\noindent\textbf{Sensitivity to Modality Fusion.}
To assess the contribution of different modalities, we conduct a sensitivity analysis by varying the geometry fusion weight \(\beta_G\) from 0 to 1. As shown in Figure~\ref{fig:para}, combining RGB and geometry evidence achieves better performance than relying on either modality alone. Specifically, UnifiedAD achieves the highest image-level AUROC of 92.8\% and pixel-level AUROC of 97.6\% at \(\beta_G=0.5\). These results demonstrate that complementary anomaly evidence from RGB and geometry modalities enables more accurate multi-modal anomaly detection.

\noindent\textbf{Shot Number Ablation.}
To evaluate the sensitivity of UnifiedAD to the number of normal reference samples, we conduct an ablation study ranging from 1-shot to 16-shot under both single-modal and multi-modal settings. As shown in Figure~\ref{fig:shot}, UnifiedAD generally benefits from additional normal references, particularly in multi-modal anomaly detection. Specifically, in the single-modal setting, image-level AUROC increases from 97.9\% to 98.3\%, while pixel-level AUROC slightly decreases from 97.8\% to 97.2\%. In the multi-modal setting, image-level and pixel-level AUROC improve from 92.8\% to 95.3\% and from 97.6\% to 99.6\%, respectively. These results demonstrate that UnifiedAD performs effectively with limited normal references while benefiting from additional samples.

\begin{table}[t]
  \centering
  \vspace{-0.3cm}
  \caption{
    \textbf{Medical Anomaly Detection on BrainAD~\cite{bao2024brainad}.}
    All values are percentages. Our results are reported as
    mean~$\pm$~standard deviation over three seeds.
    The best results are highlighted in \textbf{bold}.
  }
  \label{tab:brainad_main}

  \resizebox{\columnwidth}{!}{%
    \begin{tabular}{lcccc}
      \toprule
      \textbf{Method}
        & \textbf{I-AUROC} $\uparrow$
        & \textbf{I-AP} $\uparrow$
        & \textbf{P-AUROC} $\uparrow$
        & \textbf{P-AP} $\uparrow$ \\
      \midrule

      WinCLIP~\cite{jeong2023winclip}
        & 72.7 & 91.6 & 87.6 & 13.3 \\

      FoundAD~\cite{zhai2026foundad}
        & 76.9 & 93.9 & 95.0 & 39.1 \\

      SubspaceAD~\cite{lendering2026subspacead}
        & 78.6 & 94.0 & 96.4 & 43.9 \\

      AdaCLIP~\cite{cao2024adaclip}
        & 80.0 & 90.2 & 95.2 & 37.0 \\

      VisualAD~\cite{hou2026visualad}
        & 87.1 & 96.7 & 96.4 & 51.9 \\

      \midrule
      \rowcolor[HTML]{EAF7EB}
      \textbf{UnifiedAD (Ours)}
        & \textbf{94.0$_{\pm0.9}$}
        & \textbf{98.6$_{\pm0.2}$}
        & \textbf{96.7$_{\pm0.4}$}
        & \textbf{53.6$_{\pm3.6}$} \\

      \bottomrule
    \end{tabular}%
  }
  \vspace{-0.3cm}
\end{table}

\noindent\textbf{Cross Domain Detection.}
To assess cross-domain generalization, we evaluate UnifiedAD on BrainAD~\cite{bao2024brainad} using anomaly neurons selected from industrial data. Following BrainAD~\cite{bao2024brainad}, we report the same evaluation metrics. As shown in Table~\ref{tab:brainad_main}, UnifiedAD achieves the best results across all four reported metrics, including 94.0\% image-level AUROC and 53.6\% pixel-level AP. For more experimental details see Appendix.

\section{Conclusion}
We propose UnifiedAD, a unified framework for single-modal and multi-modal anomaly detection requiring one normal reference per target category and no gradient updates.
It combines discriminative priors from auxiliary data with transformation stability to select anomaly neurons and model their normal responses using low-rank subspaces.
Empirical-tail calibration converts projection residuals and anomaly-direction responses into comparable anomaly evidence for aggregation across layers and modalities.
Experiments on industrial benchmarks demonstrate its effectiveness in image-level anomaly detection and pixel-level anomaly localization.

\bibliographystyle{plainnat}
\bibliography{main}

\clearpage

\onecolumn
\beginappendix

The appendix is organized as follows. Appendix~\ref{app:notation} summarizes the notations used in UnifiedAD, while Appendix~\ref{app:experimental-details} provides detailed experimental settings. Appendix~\ref{More_Experiments} presents additional ablation experiments, a user study, computational efficiency analysis, and qualitative comparisons across single-modal, multi-modal, and medical anomaly detection. Appendix~\ref{app:theoretical-derivation} provides the theoretical details of evidence-guided neuron selection and auxiliary anomaly supervision. Finally, Appendices~\ref{lim} and~\ref{dis} discuss the limitations and further considerations of UnifiedAD.

\section{Notation Summary}
\label{app:notation}

For clarity, Table~\ref{tab:notation} summarizes the main notations used in UnifiedAD, covering the problem setup, evidence-guided neuron selection, one-shot calibrated subspace inference, and anomaly evidence aggregation.

\begingroup
\small
\setlength{\tabcolsep}{8pt}
\renewcommand{\arraystretch}{1.15}

\begin{longtable}{p{0.20\textwidth}p{0.44\textwidth}}
\caption{Summary of the main notations used in UnifiedAD.}
\label{tab:notation}\\

\toprule
\textbf{Notation} & \textbf{Description} \\
\midrule
\endfirsthead

\multicolumn{2}{l}{\textit{Table \thetable{} (continued)}}\\
\toprule
\textbf{Notation} & \textbf{Description} \\
\midrule
\endhead

\bottomrule
\endfoot

\multicolumn{2}{@{}l}{\textbf{Problem Setup}}\\
$c$ & Target category \\
$\mathcal{M}$ & Set of active modalities \\
$m$ & Modality index \\
$\mathcal{S}_c$ & Normal support sample of category $c$ \\
$s_c^{(m)}$ & Normal support observation of modality $m$ \\
$x^{(m)}$ & Query observation of modality $m$ \\
$y^{(m)}$ & Arbitrary input view of modality $m$ \\
$\mathcal{L}_m$ & Set of selected feature layers for modality $m$ \\
$l$ & Feature layer index \\
$\mathcal{T}_m$ & Modality-specific input adapter \\
$\Phi_l^{(m)}$ & Frozen feature encoder at layer $l$ \\
$F_l^{(m)}(y)$ & Layer-wise patch-feature matrix \\
$P_l$ & Number of spatial patches at layer $l$ \\
$D_l$ & Number of feature neurons at layer $l$ \\

\midrule
\multicolumn{2}{@{}l}{\textbf{Evidence-Guided Neuron Selection}}\\
$e_{\mathrm{aux},l,j}^{(m)}$ & Signed auxiliary anomaly effect of neuron $j$ \\
$i_{l,j}^{(m)}$ & Anomaly discriminability of neuron $j$ \\
$\tau_m$ & Normality-preserving transformation \\
$\widetilde{s}_c^{(m)}$ & Transformed normal support observation \\
$F_{l,s}^{(m)}$ & Normal support feature matrix \\
$\widetilde F_{l,s}^{(m)}$ & Spatially aligned transformed support features \\
$\eta_{l,j}^{(m)}$ & Transformation stability of neuron $j$ \\
$\epsilon_l$ & Numerical stability constant \\
$\mathcal{N}_l^{(m)}$ & Neuron-wise score normalization \\
$q_{l,j}^{(m)}$ & Unified neuron selection score \\
$w_l^{(m)}$ & Auxiliary anomaly direction \\
$S_l^{(m)}$ & Selected neuron index set \\
$K$ & Number of selected neurons (Top-$K$) \\

\midrule
\multicolumn{2}{@{}l}{\textbf{One-Shot Calibrated Subspace Inference}}\\
$X_l^{(m)}$ & Selected-neuron support feature matrix \\
$\mu_l^{(m)}$ & Neuron-wise center of support features \\
$\gamma_l^{(m)}$ & Neuron-wise scale of support features \\
$Z_l^{(m)}$ & Standardized support feature matrix \\
$U_l^{(m)}$ & Left singular vector matrix \\
$\Sigma_l^{(m)}$ & Diagonal singular value matrix \\
$V_l^{(m)}$ & Right singular vector matrix \\
$\rho_m$ & Variance-retention threshold \\
$k_l^{(m)}$ & Retained normal subspace rank \\
$B_l^{(m)}$ & Low-rank normal subspace basis \\
$F_{l,x}^{(m)}$ & Query feature matrix \\
$z_{l,p}^{(m)}$ & Standardized query patch feature \\
$\widehat z_{l,p}^{(m)}$ & Projected query patch feature \\
$r_{l,p}^{(m)}$ & Normalized subspace projection residual \\

\midrule
\multicolumn{2}{@{}l}{\textbf{Evidence Calibration and Aggregation}}\\
$\mathcal{R}_l^{(m)}$ & Calibration reference residual set \\
$\widetilde r_{l,i}^{(m)}$ & Residual of calibration patch $i$ \\
$N_l^{(m)}$ & Number of calibration reference residuals \\
$\mathbb{I}(\cdot)$ & Indicator function \\
$E_{\mathrm{sub},l,p}^{(m)}$ & Calibrated subspace anomaly evidence \\
$\bar z_{l,p}^{(m)}$ & Full-dimensional standardized query feature \\
$d_{l,p}^{(m)}$ & Auxiliary anomaly-direction response \\
$E_{\mathrm{aux},l,p}^{(m)}$ & Calibrated auxiliary-direction evidence \\
$\alpha_m$ & Auxiliary anomaly-direction evidence weight \\
$A_p^{(m)}$ & Layer-aggregated anomaly evidence for modality $m$ \\
$\beta_m$ & Modality fusion weight \\
$A_p$ & Fused patch-level anomaly evidence \\
$A$ & Pixel-level anomaly map \\
$H,W$ & Input image height and width \\
$a_{\mathrm{img}}$ & Image-level anomaly score \\

\end{longtable}
\endgroup

\section{Detailed Experimental Setup}
\label{app:experimental-details}

\subsection{Target-Support Protocol}
For MVTec 3D-AD~\cite{bergmann2021mvtec3d}, CFM~\cite{costanzino2024cfm}, CIF~\cite{lin2026cif}, M3DM~\cite{wang2023m3dm}, and CFR~\cite{guo2026cfr} are evaluated under the one-shot setting, using a single normal RGB--XYZ support pair per target category, with each baseline evaluated once. For CMDS-AD~\cite{cai2026cmds}, we use the authors' released one-shot checkpoint. UnifiedAD is evaluated using three independently sampled target supports with 3 seeds, and its results are reported as the mean~$\pm$~sample standard deviation across the three runs.

\subsection{Auxiliary Source Data and Supervision Budget}
Following the cross-dataset protocol described in the main paper, we use VisA as the auxiliary source when evaluating on MVTec-AD, MVTec 3D-AD, and medical benchmarks, while MVTec-AD serves as the auxiliary source when evaluating on VisA. The auxiliary set contains eight categories, with eight normal images and 16 anomalous images per category, yielding 64 normal images and 128 anomalous images in total. Pixel-level masks are required for the anomalous auxiliary images. These auxiliary data are used exclusively to precompute discriminative priors; no anomalous training image or pixel-level mask from MVTec 3D-AD is used.

\subsection{Subspace Retention Sensitivity Protocol}
\label{app:subspace-retention}

For the subspace retention analysis in Figure~6 of the main paper, we evaluate variance-retention thresholds of
$\{0.70, 0.80,\\ 0.90, 0.95, 0.99\}$. The horizontal-axis variable $\rho$ denotes $\rho_{\mathrm{RGB}}$ in the single-modal setting and $\rho_G$ in the multi-modal setting.
Specifically, in the multi-modal setting, we vary $\rho_G$ while keeping $\rho_{\mathrm{RGB}}=0.90$ fixed. The default thresholds are $\rho_{\mathrm{RGB}}=0.90$
and $\rho_G=0.95$. All other hyperparameters remain unchanged.

\begin{table*}[t]
  \centering
  \caption{\textbf{Medical Anomaly Detection Results on BrainAD.}
  We report image-level and pixel-level results on BrainAD~\cite{bao2024brainad}.
  All values are percentages. Our results are reported as mean~$\pm$~standard
  deviation over three seeds. The best result in each column is in \textbf{bold}.}
  \vspace{-0.2cm}
  \label{tab:medical_results}
  \resizebox{\textwidth}{!}{%
    \begin{tabular}{l ccc cccc}
      \toprule
      \multirow{2}{*}{\textbf{Method}}
        & \multicolumn{3}{c}{\textbf{Image-Level}}
        & \multicolumn{4}{c}{\textbf{Pixel-Level}} \\
      \cmidrule(lr){2-4}\cmidrule(lr){5-8}
        & I-AUROC $\uparrow$
        & I-AP $\uparrow$
        & I-$F_1$-max $\uparrow$
        & P-AUROC $\uparrow$
        & P-AP $\uparrow$
        & P-$F_1$-max $\uparrow$
        & AUPRO $\uparrow$ \\
      \midrule
      WinCLIP~\cite{jeong2023winclip}
        & 72.7 & 91.6 & 90.7 & 87.6 & 13.3 & 21.7 & 59.7 \\
      FoundAD~\cite{zhai2026foundad}
        & 76.9 & 93.9 & 90.8 & 95.0 & 39.1 & 41.7 & 76.2 \\
      SubspaceAD~\cite{lendering2026subspacead}
        & 78.6 & 94.0 & 90.9 & 96.4 & 43.9 & 50.3 & 79.5 \\
      AdaCLIP~\cite{cao2024adaclip}
        & 80.0 & 90.2 & 94.1 & 95.2 & 37.0 & 40.6 & 36.4 \\
      VisualAD~\cite{hou2026visualad}
        & 87.1 & 96.7 & 92.5 & 96.4 & 51.9 & 50.2 & \textbf{83.5} \\
      \midrule
      \rowcolor[HTML]{EAF7EB}
      \textbf{UnifiedAD (Ours)}
        & \textbf{94.0$_{\pm0.9}$}
        & \textbf{98.6$_{\pm0.2}$}
        & \textbf{94.8$_{\pm0.4}$}
        & \textbf{96.7$_{\pm0.4}$}
        & \textbf{53.6$_{\pm3.6}$}
        & \textbf{51.3$_{\pm2.6}$}
        & 83.3$_{\pm0.7}$ \\
      \bottomrule
    \end{tabular}%
  }
  \vspace{-0.3cm}
\end{table*}

\section{More Experiments}
\label{More_Experiments}

\begin{figure}[t]
\centering
\includegraphics[width=0.85\linewidth]{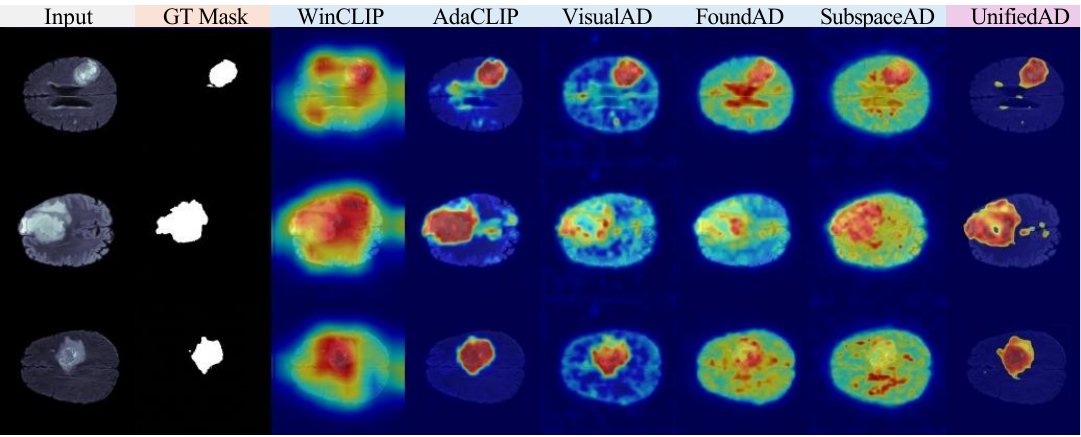}
\caption{\textbf{Qualitative Comparison on BrainAD.} Comparison of anomaly localization results across different methods.}
\label{fig:brainad}
\vspace{-0.5cm}
\end{figure}

\subsection{Medical Anomaly Detection}
To evaluate the cross-domain generalization capability of UnifiedAD, we conduct experiments on the BrainAD~\cite{bao2024brainad} dataset. Each of the three runs uses a different target-support seed to independently fit its corresponding neuron selection. As shown in Table~\ref{tab:medical_results}, UnifiedAD achieves strong performance on BrainAD without medical-domain anomaly supervision. Specifically, UnifiedAD achieves an image-level AUROC of 94.0\% and a pixel-level AP of 53.6\% on BrainAD. Furthermore, Figure~\ref{fig:brainad} shows that UnifiedAD produces more concentrated anomaly responses that closely align with the ground-truth masks, while effectively suppressing irrelevant activations in normal brain regions compared with existing methods. These results demonstrate that the anomaly neurons selected from VisA exhibit cross-domain transferability.

\subsection{Impact of Auxiliary Anomaly Supervision}
To evaluate the contribution of auxiliary anomalous samples, we conduct an ablation study on MVTec 3D-AD~\cite{bergmann2021mvtec3d} by removing auxiliary anomaly supervision while keeping all other settings fixed. Unlike B3 in the main paper, which removes only auxiliary-direction
evidence, this setting removes auxiliary anomaly supervision from
both neuron selection and direction construction. Specifically,
RGB neurons are selected solely based on transformation stability
by setting $i_{l,j}^{(\mathrm{RGB})}=1$ in Eq.~(3) in the main paper. As shown in Table~\ref{tab:auxiliary_anomaly_ablation}, auxiliary anomalous samples significantly improve image-level anomaly detection and pixel-level anomaly localization. Specifically, removing auxiliary anomaly supervision decreases image-level AUROC by 10.2 percentage points and pixel-level AP and $F_1$-max by 12.9 and 9.7 points, respectively, while pixel-level AUROC increases by 1.2 points. These results demonstrate that auxiliary anomaly supervision provides discriminative anomaly-direction evidence that complements normal subspace modeling, improving anomaly discrimination and pixel-level localization precision while maintaining strong overall localization performance.

\begin{table*}[h]
  \centering
  \caption{\textbf{Effect of Removing Auxiliary Anomalous Samples on MVTec 3D-AD.}
  All values are percentages and are reported as mean~$\pm$~sample standard deviation over support seeds 23, 29, and 37.
  Gain is Full UnifiedAD minus the setting without auxiliary anomalous samples, in percentage points (pp); a negative value indicates a decrease. The best result in each metric column is in \textbf{bold}.}
  \label{tab:auxiliary_anomaly_ablation}
  \resizebox{0.98\textwidth}{!}{%
    \begin{tabular}{lccccccc}
      \toprule
      \multirow{2}{*}{\textbf{Setting}}
        & \multicolumn{3}{c}{\textbf{Image-Level}}
        & \multicolumn{4}{c}{\textbf{Pixel-Level}} \\
      \cmidrule(lr){2-4}\cmidrule(lr){5-8}
        & AUROC $\uparrow$ & AP $\uparrow$ & $F_1$-max $\uparrow$
        & AUROC $\uparrow$ & AP $\uparrow$ & $F_1$-max $\uparrow$ & AUPRO $\uparrow$ \\
      \midrule
      Without Auxiliary Anomalies
        & 82.6$_{\pm 2.5}$
        & 94.4$_{\pm 1.1}$
        & 93.0$_{\pm 0.6}$
        & \textbf{98.8$_{\pm 0.1}$}
        & 49.6$_{\pm 3.0}$
        & 50.7$_{\pm 2.4}$
        & 96.1$_{\pm 0.3}$ \\
      \rowcolor[HTML]{EAF7EB}
      \textbf{Full UnifiedAD (Ours)}
        & \textbf{92.8$_{\pm 0.3}$}
        & \textbf{97.7$_{\pm 0.1}$}
        & \textbf{93.9$_{\pm 0.1}$}
        & 97.6$_{\pm 0.1}$
        & \textbf{62.5$_{\pm 0.9}$}
        & \textbf{60.4$_{\pm 1.0}$}
        & \textbf{97.5$_{\pm 0.1}$} \\
      \midrule
      \textbf{Gain (pp)}
        & +10.2 & +3.3 & +0.9 & $-1.2$ & +12.9 & +9.7 & +1.4 \\
      \bottomrule
    \end{tabular}%
  }
\end{table*}

\subsection{Calibration-View Transformation Analysis}
\begin{figure}[h]
\centering
\includegraphics[width=0.99\linewidth]{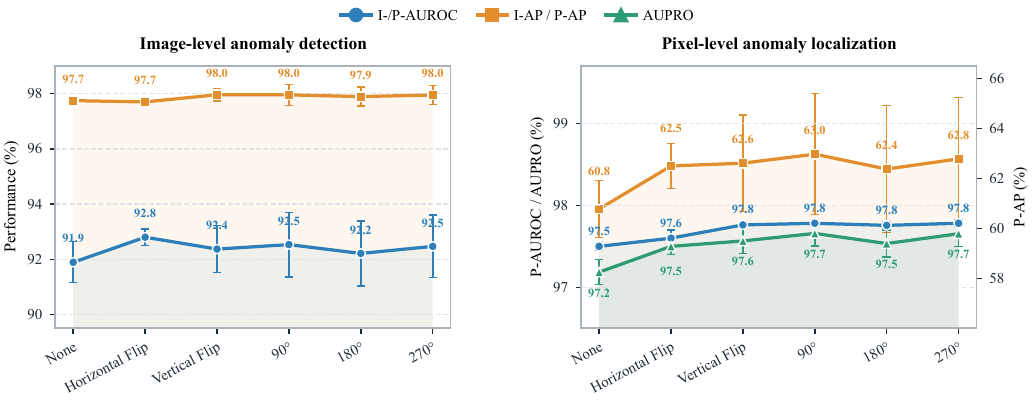}
\vspace{-0.3cm}
\caption{ \textbf{Calibration-View Transformation Analysis.}
    Comparison of no transformation, horizontal and vertical flipping, and rotations of 90°, 180°, and 270° applied to the normal reference for empirical-tail calibration on MVTec 3D-AD. Image-level anomaly detection results are shown on the left, and pixel-level anomaly localization results on the right. Horizontal flipping is the default setting in UnifiedAD. All values are percentages.}
\label{fig:flip}
\vspace{-0.5cm}
\end{figure}

To investigate whether the choice of calibration-view transformation affects anomaly detection, we compare different flips and rotations of the normal reference with no transformation. As shown in Figure~\ref{fig:flip}, every transformed view improves I-AUROC, P-AP, and AUPRO over the untransformed reference. Specifically, horizontal flipping increases I-AUROC from 91.9\% to 92.8\%, P-AP from 60.8\% to 62.5\%, and AUPRO from 97.2\% to 97.5\%. The other transformations achieve similar results, with I-AUROC ranging from 92.2\% to 92.5\% and P-AP from 62.4\% to 63.0\%. These results show that transformed normal references provide useful variation for empirical-tail calibration. Horizontal flipping is used by default because it achieves the highest I-AUROC among the evaluated settings while maintaining competitive pixel-level performance.

\subsection{Backbone Generalizability in Single-Modal Anomaly Detection}
To investigate whether UnifiedAD generalizes across visual backbones in single-modal anomaly detection, we evaluate it with CLIP ViT-L/14 (CLIP-L)~\cite{radford2021learning}, DINOv3 ViT-L/16 (DINOv3-L)~\cite{simeoni2025dinov3}, and DINOv2 ViT-g/14 (DINOv2-G)~\cite{oquab2023dinov2} on MVTec-AD~\cite{bergmann2019mvtec} and VisA~\cite{zou2022spot}. As shown in Table~\ref{tab:backbone_generalizability_mvtec_visa}, UnifiedAD can be applied with all three backbones, while DINOv2-G achieves the best result in every applicable metric on both datasets. Specifically, DINOv2-G attains image-level AUROC scores of 97.9\% and 96.3\%, and pixel-level AP scores of 68.4\% and 52.2\% on MVTec-AD and VisA, respectively. DINOv3-L improves pixel-level AP over CLIP on both datasets, although CLIP achieves slightly higher image-level AUROC. These results demonstrate that UnifiedAD generalizes across backbone families, with the choice of backbone affecting detection and localization differently.

\begin{table*}[h]
  \centering
  \caption{
    \textbf{Backbone Generalizability of UnifiedAD on MVTec-AD and VisA.}
    We evaluate UnifiedAD with three visual backbones:
    CLIP-L~\cite{radford2021learning},
    DINOv2-G~\cite{oquab2023dinov2},
    and DINOv3-L~\cite{simeoni2025dinov3}.
    All results are reported as mean~$\pm$~standard deviation.
    The best result in each column is highlighted in \textbf{bold}.
    N/A denotes metrics that are not applicable.
  }
  \vspace{-0.2cm}
  \label{tab:backbone_generalizability_mvtec_visa}
  \resizebox{0.98\textwidth}{!}{%
    \begin{tabular}{clcccccccc}
      \toprule
      \multirow{2}{*}{\textbf{Level}}
        & \multirow{2}{*}{\textbf{Backbone}}
        & \multicolumn{4}{c}{\textbf{MVTec-AD~\cite{bergmann2019mvtec}}}
        & \multicolumn{4}{c}{\textbf{VisA~\cite{zou2022spot}}} \\
      \cmidrule(lr){3-6}\cmidrule(lr){7-10}
        &
        & AUROC $\uparrow$
        & AP $\uparrow$
        & $F_1$-max $\uparrow$
        & AUPRO $\uparrow$
        & AUROC $\uparrow$
        & AP $\uparrow$
        & $F_1$-max $\uparrow$
        & AUPRO $\uparrow$ \\
      \midrule

      \multirow{3}{*}{Image}
        & \cellcolor[HTML]{EAF7EB}UnifiedAD (CLIP-L~\cite{radford2021learning})
          & \cellcolor[HTML]{EAF7EB}94.8$_{\pm 0.3}$
          & \cellcolor[HTML]{EAF7EB}97.6$_{\pm 0.1}$
          & \cellcolor[HTML]{EAF7EB}95.0$_{\pm 0.5}$
          & \cellcolor[HTML]{EAF7EB}\textcolor{gray}{N/A}
          & \cellcolor[HTML]{EAF7EB}90.6$_{\pm 1.5}$
          & \cellcolor[HTML]{EAF7EB}92.2$_{\pm 1.0}$
          & \cellcolor[HTML]{EAF7EB}86.7$_{\pm 0.7}$
          & \cellcolor[HTML]{EAF7EB}\textcolor{gray}{N/A} \\

        & \cellcolor[HTML]{EAF7EB}UnifiedAD (DINOv3-L~\cite{simeoni2025dinov3})
          & \cellcolor[HTML]{EAF7EB}93.5$_{\pm 0.6}$
          & \cellcolor[HTML]{EAF7EB}96.5$_{\pm 0.2}$
          & \cellcolor[HTML]{EAF7EB}94.4$_{\pm 0.3}$
          & \cellcolor[HTML]{EAF7EB}\textcolor{gray}{N/A}
          & \cellcolor[HTML]{EAF7EB}89.7$_{\pm 2.2}$
          & \cellcolor[HTML]{EAF7EB}90.9$_{\pm 1.8}$
          & \cellcolor[HTML]{EAF7EB}87.1$_{\pm 0.7}$
          & \cellcolor[HTML]{EAF7EB}\textcolor{gray}{N/A} \\
          & \cellcolor[HTML]{EAF7EB}\textbf{UnifiedAD (DINOv2-G}~\cite{oquab2023dinov2}\textbf{)}
  & \cellcolor[HTML]{EAF7EB}\textbf{97.9$_{\pm 0.2}$}
  & \cellcolor[HTML]{EAF7EB}\textbf{99.0$_{\pm 0.1}$}
  & \cellcolor[HTML]{EAF7EB}\textbf{96.9$_{\pm 0.1}$}
  & \cellcolor[HTML]{EAF7EB}\textcolor{gray}{N/A}
  & \cellcolor[HTML]{EAF7EB}\textbf{96.3$_{\pm 0.1}$}
  & \cellcolor[HTML]{EAF7EB}\textbf{96.8$_{\pm 0.1}$}
  & \cellcolor[HTML]{EAF7EB}\textbf{92.9$_{\pm 0.1}$}
  & \cellcolor[HTML]{EAF7EB}\textcolor{gray}{N/A} \\

      \midrule

      \multirow{3}{*}{Pixel}
        & \cellcolor[HTML]{EAF7EB}UnifiedAD (CLIP-L~\cite{radford2021learning})
          & \cellcolor[HTML]{EAF7EB}95.4$_{\pm 0.1}$
          & \cellcolor[HTML]{EAF7EB}52.7$_{\pm 0.5}$
          & \cellcolor[HTML]{EAF7EB}53.5$_{\pm 0.5}$
          & \cellcolor[HTML]{EAF7EB}91.3$_{\pm 0.3}$
          & \cellcolor[HTML]{EAF7EB}90.8$_{\pm 0.1}$
          & \cellcolor[HTML]{EAF7EB}17.7$_{\pm 0.1}$
          & \cellcolor[HTML]{EAF7EB}25.5$_{\pm 0.0}$
          & \cellcolor[HTML]{EAF7EB}74.6$_{\pm 0.3}$ \\

        & \cellcolor[HTML]{EAF7EB}UnifiedAD (DINOv3-L~\cite{simeoni2025dinov3})
          & \cellcolor[HTML]{EAF7EB}96.9$_{\pm 0.1}$
          & \cellcolor[HTML]{EAF7EB}61.3$_{\pm 0.5}$
          & \cellcolor[HTML]{EAF7EB}59.8$_{\pm 0.5}$
          & \cellcolor[HTML]{EAF7EB}93.1$_{\pm 0.1}$
          & \cellcolor[HTML]{EAF7EB}97.7$_{\pm 0.1}$
          & \cellcolor[HTML]{EAF7EB}37.9$_{\pm 0.7}$
          & \cellcolor[HTML]{EAF7EB}42.9$_{\pm 0.5}$
          & \cellcolor[HTML]{EAF7EB}92.8$_{\pm 0.2}$ \\

& \cellcolor[HTML]{EAF7EB}\textbf{UnifiedAD (DINOv2-G}~\cite{oquab2023dinov2}\textbf{)}
  & \cellcolor[HTML]{EAF7EB}\textbf{97.8$_{\pm 0.1}$}
  & \cellcolor[HTML]{EAF7EB}\textbf{68.4$_{\pm 0.5}$}
  & \cellcolor[HTML]{EAF7EB}\textbf{64.9$_{\pm 0.6}$}
  & \cellcolor[HTML]{EAF7EB}\textbf{94.7$_{\pm 0.1}$}
  & \cellcolor[HTML]{EAF7EB}\textbf{99.2$_{\pm 0.0}$}
  & \cellcolor[HTML]{EAF7EB}\textbf{52.2$_{\pm 0.8}$}
  & \cellcolor[HTML]{EAF7EB}\textbf{54.2$_{\pm 0.5}$}
  & \cellcolor[HTML]{EAF7EB}\textbf{95.1$_{\pm 0.2}$} \\
        \bottomrule
    \end{tabular}%
  }
\end{table*}

\subsection{Backbone Generalizability in Multi-Modal Anomaly Detection}
To investigate whether UnifiedAD generalizes across visual backbones, we evaluate it with CLIP-L~\cite{radford2021learning}, DINOv3-L~\cite{simeoni2025dinov3}, and DINOv2-G~\cite{oquab2023dinov2} on MVTec 3D-AD. As shown in Table~\ref{tab:mvtec3d_multimodal_results}, UnifiedAD produces anomaly detection and localization results with all three backbones, while DINOv2-G achieves the best result in six of the seven metrics. Specifically, image-level AUROC increases from 83.9\% with CLIP to 89.3\% with DINOv3-L and 92.8\% with DINOv2-G; pixel-level AP increases from 38.8\% to 56.4\% and 62.5\%, respectively. DINOv3-L achieves the highest pixel-level AUROC at 98.9\%. These results demonstrate that UnifiedAD can be applied across different backbone families, although backbone choice affects its performance.

\begin{table*}[h]
  \centering
  \caption{\textbf{Backbone Generalizability of UnifiedAD on MVTec 3D-AD.}
  We evaluate UnifiedAD with three different visual backbones:
  CLIP-L~\cite{radford2021learning}, DINOv2-G~\cite{oquab2023dinov2}, and DINOv3-L~\cite{simeoni2025dinov3}.
  Results are reported as mean~$\pm$~sample standard deviation
  over three support seeds. All values are percentages.
  The best result in each column is highlighted in \textbf{bold}.}
  \vspace{-0.2cm}
  \label{tab:mvtec3d_multimodal_results}
  \resizebox{0.98\textwidth}{!}{%
    \setlength{\tabcolsep}{12pt}%
    \begin{tabular}{lccccccc}
      \toprule
      \multirow{2}{*}{\textbf{Method}}
        & \multicolumn{3}{c}{\textbf{Image-Level}}
        & \multicolumn{4}{c}{\textbf{Pixel-Level}} \\
      \cmidrule(lr){2-4}\cmidrule(lr){5-8}
        & AUROC $\uparrow$ & AP $\uparrow$ & $F_1$-max $\uparrow$
        & AUROC $\uparrow$ & AP $\uparrow$ & $F_1$-max $\uparrow$ & AUPRO $\uparrow$ \\
      \midrule
      \cellcolor[HTML]{EAF7EB}UnifiedAD (CLIP-L~\cite{radford2021learning})
  & \cellcolor[HTML]{EAF7EB}83.9$_{\pm 1.5}$
  & \cellcolor[HTML]{EAF7EB}95.2$_{\pm 0.5}$
  & \cellcolor[HTML]{EAF7EB}91.1$_{\pm 0.2}$
  & \cellcolor[HTML]{EAF7EB}98.0$_{\pm 0.0}$
  & \cellcolor[HTML]{EAF7EB}38.8$_{\pm 0.4}$
  & \cellcolor[HTML]{EAF7EB}42.9$_{\pm 0.3}$
  & \cellcolor[HTML]{EAF7EB}93.2$_{\pm 0.2}$ \\

\cellcolor[HTML]{EAF7EB}UnifiedAD (DINOv3-L~\cite{simeoni2025dinov3})
  & \cellcolor[HTML]{EAF7EB}89.3$_{\pm 0.6}$
  & \cellcolor[HTML]{EAF7EB}97.0$_{\pm 0.2}$
  & \cellcolor[HTML]{EAF7EB}93.3$_{\pm 0.3}$
  & \cellcolor[HTML]{EAF7EB}\textbf{98.9$_{\pm 0.0}$}
  & \cellcolor[HTML]{EAF7EB}56.4$_{\pm 1.5}$
  & \cellcolor[HTML]{EAF7EB}56.5$_{\pm 0.9}$
  & \cellcolor[HTML]{EAF7EB}96.6$_{\pm 0.1}$ \\
      \cellcolor[HTML]{EAF7EB}\textbf{UnifiedAD (DINOv2-G}~\cite{oquab2023dinov2}\textbf{)}
        & \cellcolor[HTML]{EAF7EB}\textbf{92.8$_{\pm 0.3}$}
        & \cellcolor[HTML]{EAF7EB}\textbf{97.7$_{\pm 0.1}$}
        & \cellcolor[HTML]{EAF7EB}\textbf{93.9$_{\pm 0.1}$}
        & \cellcolor[HTML]{EAF7EB}97.6$_{\pm 0.1}$
        & \cellcolor[HTML]{EAF7EB}\textbf{62.5$_{\pm 0.9}$}
        & \cellcolor[HTML]{EAF7EB}\textbf{60.4$_{\pm 1.0}$}
        & \cellcolor[HTML]{EAF7EB}\textbf{97.5$_{\pm 0.1}$} \\
      \bottomrule
    \end{tabular}%
  }
  \vspace{-0.3cm}
\end{table*}

\subsection{Sensitivity to Top-K Neuron Selection}
\begin{figure}[h]
\centering
\includegraphics[width=0.5\linewidth]{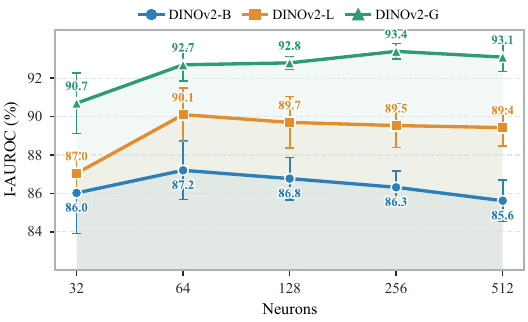}
\vspace{-0.3cm}
\caption{\textbf{Top-$K$ Ablation on MVTec 3D-AD.} I-AUROC across different numbers of selected neurons for DINOv2-B/L/G.}
\label{fig:k}
\vspace{-0.5cm}
\end{figure}

To examine the sensitivity of UnifiedAD to the number of selected neurons across backbone scales, we vary \(K\) from 32 to 512 on MVTec 3D-AD using DINOv2 ViT-b/14, ViT-l/14, and ViT-g/14. As shown in Figure~\ref{fig:k}, increasing \(K\) from 32 to 64 improves I-AUROC for all three backbones: from 86.0\% to 87.2\% for DINOv2-B, from 87.0\% to 90.1\% for DINOv2-L, and from 90.7\% to 92.7\% for DINOv2-G. Beyond \(K=64\), DINOv2-B and DINOv2-L show no further improvement, whereas DINOv2-G reaches its highest observed I-AUROC of 93.4\% at \(K=256\). Increasing \(K\) to 512 reduces performance for all three backbones. DINOv2-G consistently achieves the highest I-AUROC across the evaluated values of \(K\). At \(K=128\), each backbone remains within 0.6 percentage points of its respective peak, supporting \(K=128\) as a stable shared setting without backbone-specific tuning.

\subsection{User Study}
\begin{figure}[h]
    \centering
    \begin{subfigure}{0.49\linewidth}
        \centering
        \includegraphics[width=\linewidth]{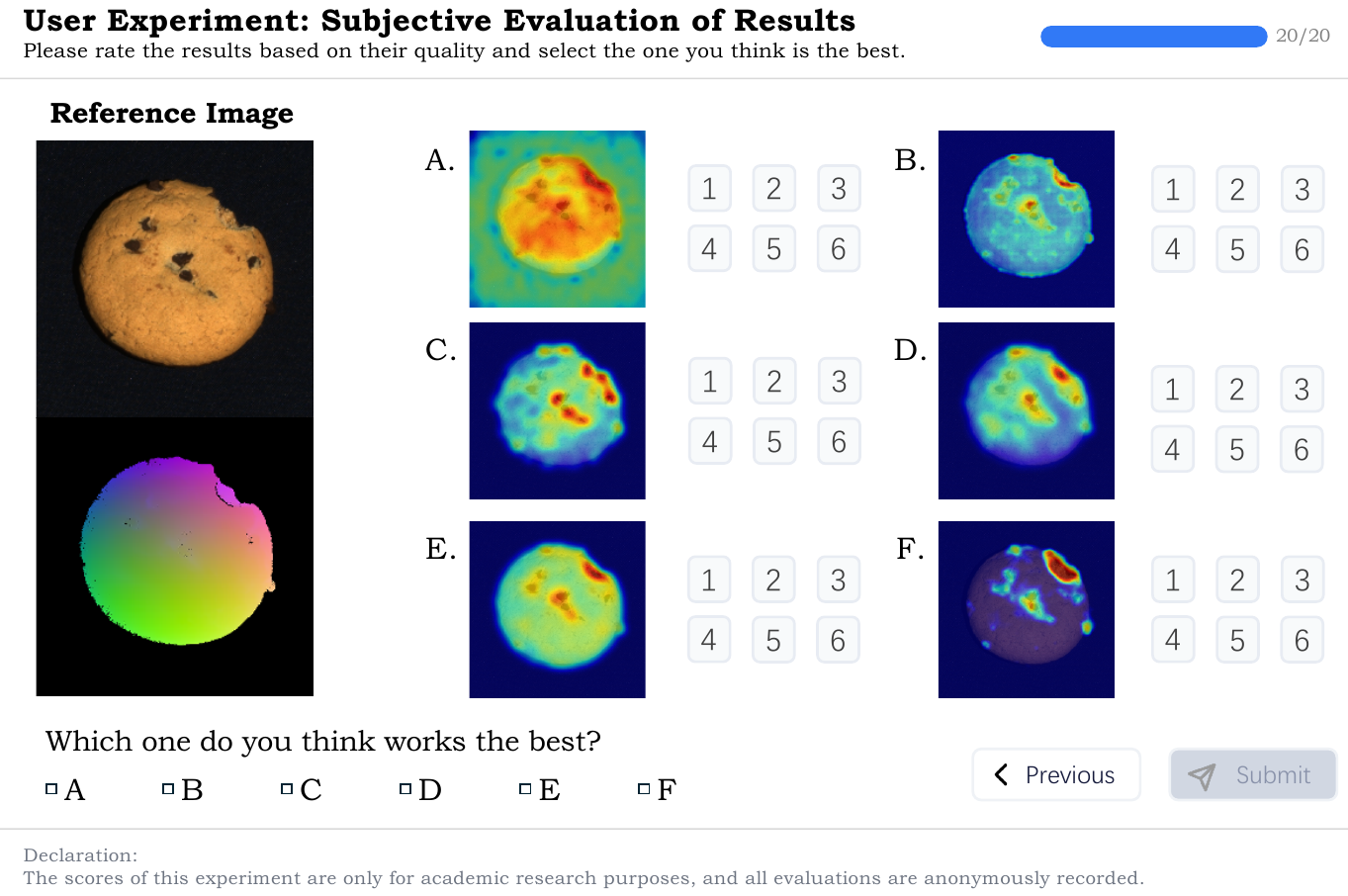}
        \caption{}
        \label{fig:user}
    \end{subfigure}
    \hfill
    \begin{subfigure}{0.49\linewidth}
        \centering
        \includegraphics[width=\linewidth]{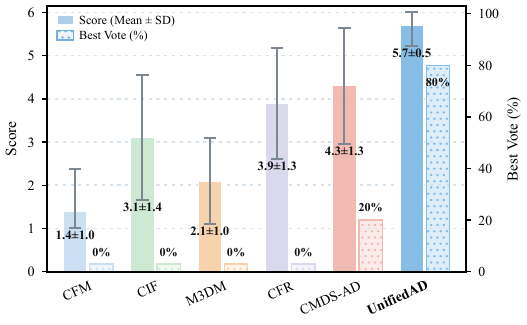}
        \caption{}
        \label{fig:user_survey}
    \end{subfigure}
    \caption{\textbf{User Study.} Overview of the subjective evaluation interface and user survey.}
    \label{fig:user_study}
\end{figure}

To evaluate the perceptual quality of anomaly localization, we conduct a user study with 10 participants (5 male and 5 female) from diverse academic and professional backgrounds. As shown in Figure~\ref{fig:user_study}(a), participants independently rate the anonymized and randomly shuffled results of six methods and select the best result for each sample. Figure~\ref{fig:user_study}(b) shows that UnifiedAD achieves the highest mean score of 5.7, outperforming the strongest baseline, CMDS-AD (4.3), by 1.4 points. Specifically, UnifiedAD receives 80\% of the best-result votes, while CMDS-AD receives the remaining 20\%, with no votes for other methods. These results demonstrate that UnifiedAD achieves superior subjective evaluation scores and is consistently preferred for anomaly localization across the evaluated samples.

\subsection{Computational Efficiency Analysis}
To evaluate the computational efficiency of UnifiedAD, we measure inference latency, throughput, peak GPU memory, and one-time adaptation cost under different configurations. As shown in Table~\ref{tab:efficiency_analysis}, UnifiedAD maintains the same peak GPU memory usage as the frozen DINOv2-G backbone while introducing moderate computational overhead. Specifically, the single-modal configuration achieves a latency of 146.8 ms and a throughput of 6.81 FPS, while the multi-modal end-to-end configuration requires 342.1 ms per sample and achieves 2.94 FPS. Meanwhile, the one-time adaptation costs are only approximately 0.40 s and 1.57 s for single-modal and multi-modal settings, respectively. These results demonstrate that UnifiedAD enables efficient adaptation to new categories without gradient-based training, while maintaining a stable GPU memory footprint across different input settings.

\begin{table*}[h]
  \centering
  \caption{
    \textbf{Computational Efficiency Analysis.}
    Comparison of inference latency, throughput, peak GPU memory,
    and one-time adaptation cost under different configurations.
    Latency is measured in milliseconds (ms), throughput in frames
    per second (FPS), memory in GiB, and adaptation time in seconds (s).
    UnifiedAD configurations are highlighted in light green.
  }
  \label{tab:efficiency_analysis}

  \resizebox{0.98\textwidth}{!}{%
    \begin{tabular}{lcccc}
      \toprule
      \textbf{Configuration}
        & Latency (ms) $\downarrow$
        & Throughput (FPS) $\uparrow$
        & Peak Memory (GiB) $\downarrow$
        & One-Time Adaptation (s) $\downarrow$ \\
      \midrule

      DINOv2-G (Backbone Only)
        & 98.7$_{\pm 3.6}$
        & 10.13
        & 4.64
        & \textcolor{gray}{---} \\

      \rowcolor[HTML]{EAF7EB}
      \textbf{UnifiedAD (Single-Modal)}
        & 146.8$_{\pm 9.9}$
        & 6.81
        & 4.64
        & $\approx$0.40 \\

      \rowcolor[HTML]{EAF7EB}
      \textbf{UnifiedAD (Multi-Modal Core)}
        & 252.8$_{\pm 11.7}$
        & 3.95
        & 4.64
        & 1.57 \\

      \rowcolor[HTML]{EAF7EB}
      \textbf{UnifiedAD (Multi-Modal End-to-End)}
        & 342.1$_{\pm 27.9}$
        & 2.94
        & 4.64
        & 1.57 \\

      \bottomrule
    \end{tabular}%
  }
\end{table*}

\subsection{More Qualitative Comparison on Single-Modal Anomaly Detection}
\begin{figure*}[h]
\centering
\includegraphics[width=\linewidth]{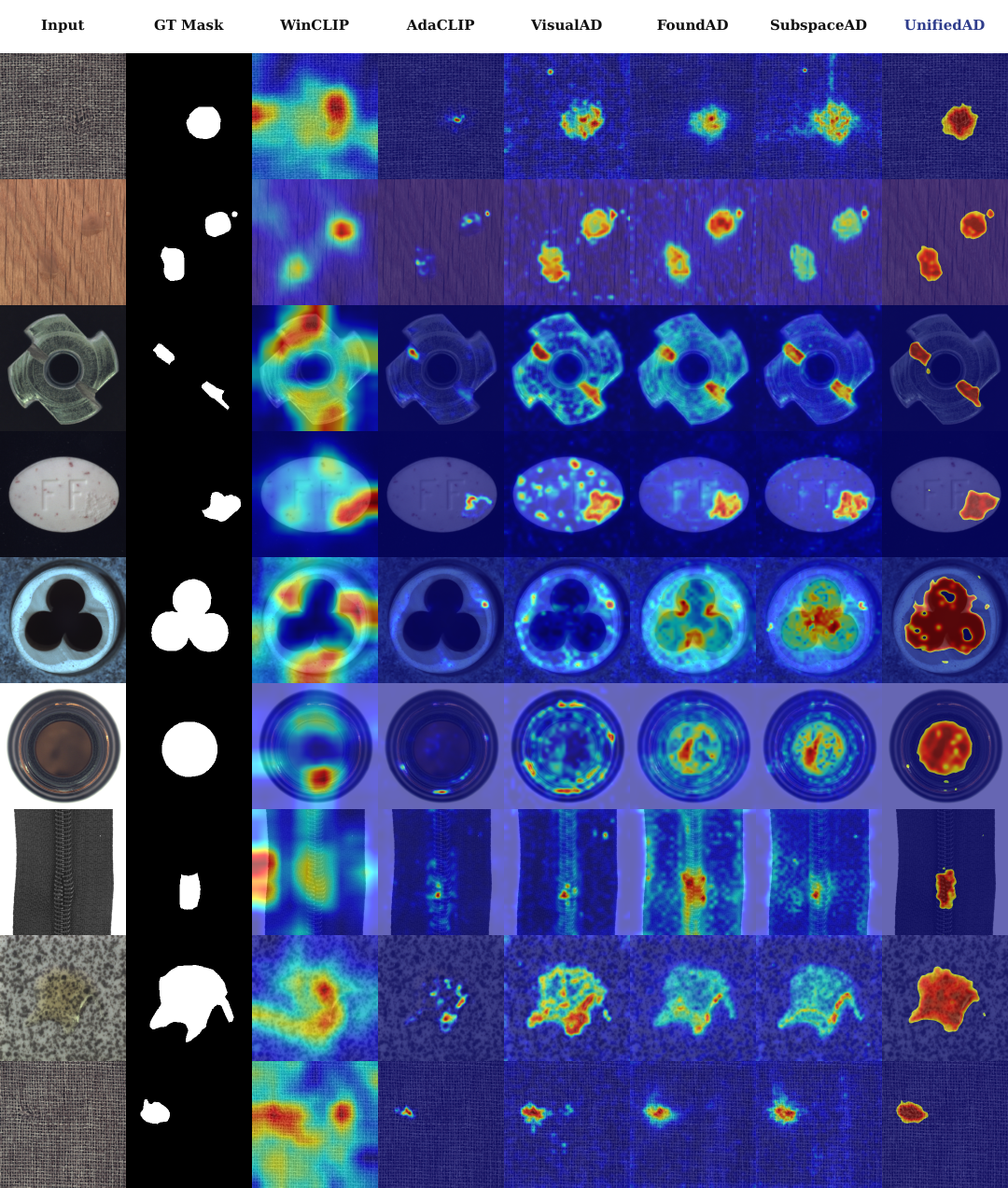}
\caption{\textbf{More Qualitative Comparison on MVTec-AD.} Comparison of anomaly localization results across different single-modal methods.}
\label{fig:mvtec_more}
\end{figure*}

To further evaluate the anomaly localization capability of UnifiedAD in single-modal settings, we provide additional qualitative comparisons on the MVTec-AD~\cite{bergmann2019mvtec} benchmark. As shown in Figure~\ref{fig:mvtec_more}, existing methods often produce diffuse anomaly responses or false activations in normal regions, leading to inaccurate defect localization. In contrast, UnifiedAD generates more concentrated anomaly responses that closely align with the ground-truth masks, even for defects with different shapes, sizes, and textures. These results further demonstrate the effectiveness of UnifiedAD in distinguishing anomalous deviations from normal variations, enabling more precise and robust anomaly localization across diverse defect types.

\subsection{More Qualitative Comparison on Multi-Modal Anomaly Detection}

\begin{figure*}[h]
\centering
\includegraphics[width=\linewidth]{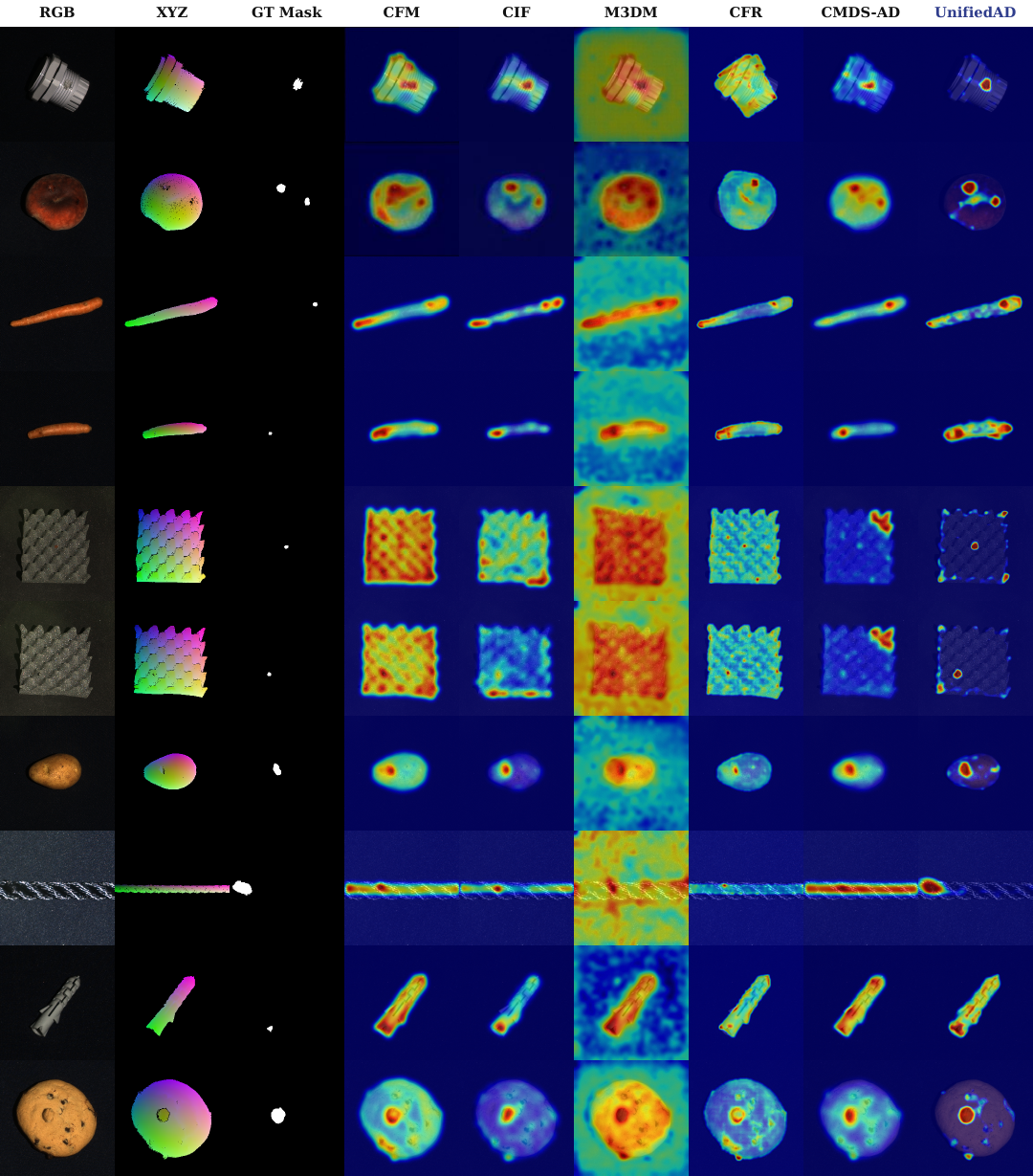}
\caption{\textbf{More Qualitative Comparison on MVTec 3D-AD.}
Comparison of anomaly localization results across different multi-modal methods. UnifiedAD produces more concentrated anomaly responses that better align with the ground-truth masks while suppressing irrelevant activations in normal regions.}
\label{fig:mvtec3d_more}
\end{figure*}

To further evaluate the anomaly localization capability of UnifiedAD in multi-modal settings, we provide additional qualitative comparisons on the MVTec 3D-AD benchmark. As shown in Figure~\ref{fig:mvtec3d_more}, existing multi-modal methods often produce diffuse anomaly responses or introduce irrelevant activations in normal regions, which reduces localization accuracy. In contrast, UnifiedAD generates more concentrated anomaly responses that better align with the ground-truth masks across diverse categories and anomaly patterns. These results further demonstrate that UnifiedAD can effectively integrate complementary anomaly evidence from RGB and geometry modalities, enabling more accurate and robust multi-modal anomaly localization.

\subsection{More Qualitative Comparison on Medical Anomaly Detection}

\begin{figure*}[h]
\centering
\includegraphics[width=\linewidth]{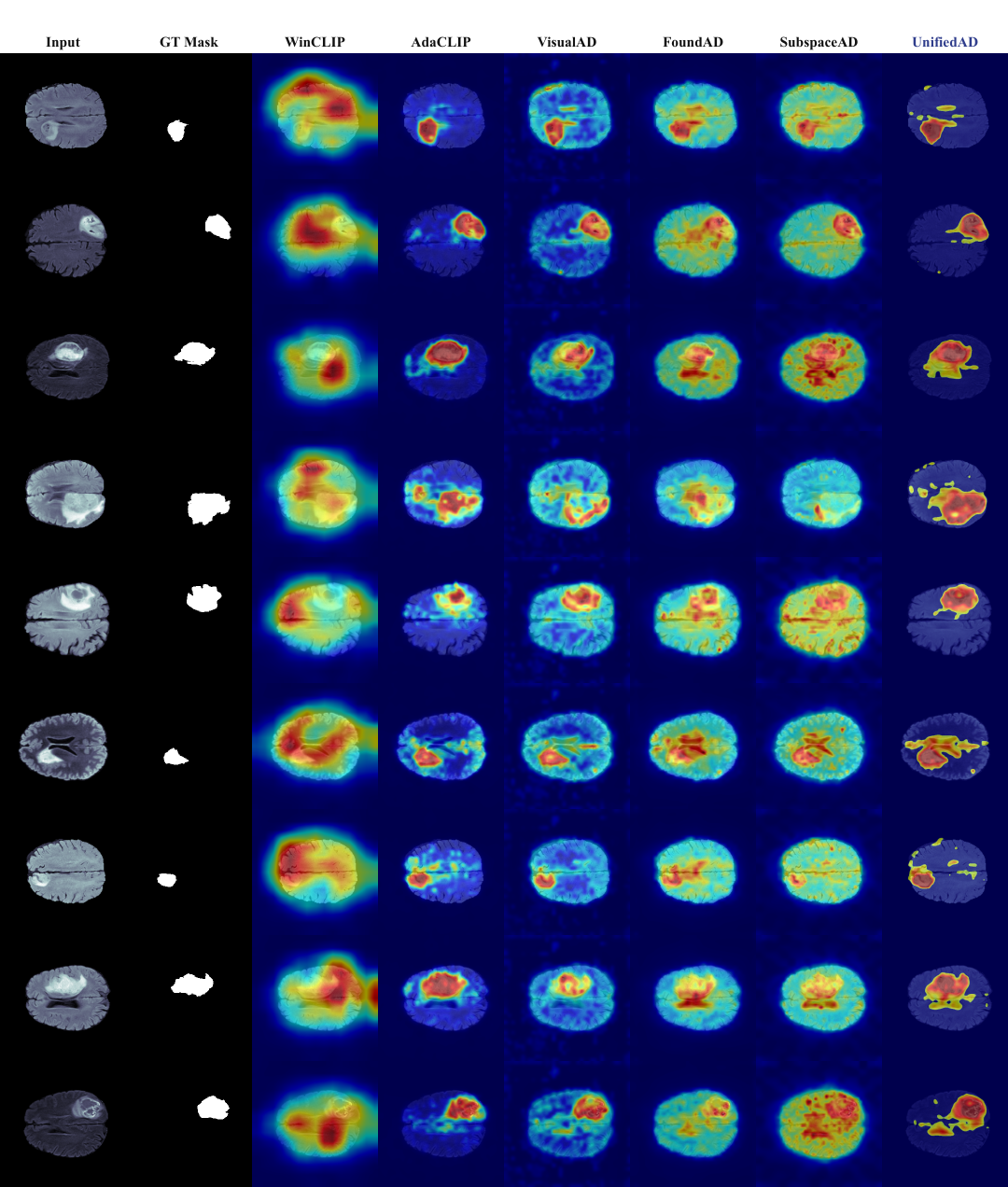}
\caption{\textbf{More Qualitative Comparison on BrainAD.} Comparison of anomaly localization results across different methods. UnifiedAD produces more concentrated anomaly responses that closely align with the ground-truth masks while suppressing irrelevant activations in normal brain regions.}
\label{fig:brainad_more}
\end{figure*}

To further evaluate the anomaly localization capability of UnifiedAD in medical settings, we provide additional qualitative comparisons on the BrainAD~\cite{bao2024brainad} benchmark. As shown in Figure~\ref{fig:brainad_more}, existing methods often produce diffuse anomaly responses or false activations in normal brain regions, making it difficult to precisely localize abnormal regions. In contrast, UnifiedAD generates more concentrated anomaly responses that better align with the ground-truth masks across different brain images and anomaly patterns. These results further demonstrate the effectiveness of UnifiedAD in distinguishing anomalous deviations from normal variations, enabling more accurate and reliable anomaly localization in medical images.

\section{Theoretical Derivation}
\label{app:theoretical-derivation}

\subsection{Signed Anomaly Effect and Supervision Requirements}
For each auxiliary category $a$, feature layer $l$, and neuron $j$, we partition the auxiliary patches into three sets: normal-image patches ($N$), patches inside the anomaly mask ($I$), and background patches outside the anomaly mask in anomalous images ($O$). We first measure the activation shift from normal to anomalous regions:
\begin{equation}
\delta_{a,l,j}
=
\frac{\mu_{a,l,j}^{I}-\mu_{a,l,j}^{N}}
{\sqrt{v_{a,l,j}^{N}+\epsilon}},
\end{equation}
where $\mu^{N}_{a,l,j}$ and $v^{N}_{a,l,j}$ denote the mean and variance of the normal-patch activations. We additionally measure the localization contrast between the anomalous region and the background of the same anomalous images:
\begin{equation}
\lambda_{a,l,j}
=
\frac{\mu_{a,l,j}^{I}-\mu_{a,l,j}^{O}}
{\sqrt{\tfrac{1}{2}\left(v_{a,l,j}^{I}+v_{a,l,j}^{O}\right)+\epsilon}}.
\end{equation}
A neuron is retained at the category level only when these two effects have consistent directions. Its signed category-specific anomaly effect is
\begin{equation}
e_{a,l,j}
=
\operatorname{sgn}\!\left(\delta_{a,l,j}\right)
\sqrt{\left|\delta_{a,l,j}\lambda_{a,l,j}\right|}
\mathbb{I}\!\left[\delta_{a,l,j}\lambda_{a,l,j}>0\right].
\end{equation}
Across auxiliary categories, we determine the final effect $e_{\mathrm{aux},l,j}$ using the median response magnitude and the dominant direction among categories, subject to the cross-category agreement threshold $\tau_{\mathrm{agr}}$. This effect is subsequently combined with horizontal-flip stability to select the top-$K$ neurons and construct a sparse signed anomaly direction.

This computation requires pixel-level anomaly masks only for the auxiliary source-domain data. It does not use anomalous images or pixel-level annotations from the target dataset. In particular, the geometry branch on MVTec 3D-AD does not use labeled geometric anomalies; its neurons are selected solely from the normal RGB--XYZ support and its spatially aligned horizontal flip.

\section{Limitation}
\label{lim}

Although UnifiedAD demonstrates strong performance across diverse anomaly detection benchmarks, some practical limitations remain. Specifically, the spatial resolution of the generated anomaly maps is constrained by the patch-level representations of the frozen visual encoder, which may affect the localization of extremely small defects. Future work will explore finer-grained anomaly localization to further improve the flexibility of UnifiedAD.

\section{More Discussion}
\label{dis}

\noindent$\triangleright$ \textbf{\textit{Q1. What Is Unified Across Different Input Modalities?}}

A key insight of UnifiedAD is that anomaly detection across different input modalities can be formulated through a shared normality modeling and evidence assessment mechanism. Although visual appearance and geometric structure exhibit different feature characteristics, anomalous observations can be characterized by their deviations from the normal response structures of pretrained representations. This motivates the use of modality-specific normal subspaces under a common projection-residual formulation. By further calibrating these deviations against normal reference variations, UnifiedAD establishes a consistent form of anomaly evidence across feature layers and modalities, allowing heterogeneous information to be aggregated within the same inference framework. In this way, unification is achieved at the level of normality characterization and anomaly evidence assessment, while preserving the distinct representations and complementary information provided by individual modalities.

\noindent$\triangleright$ \textbf{\textit{Q2. How Does Transferable Anomaly Knowledge Support One-Shot Normality Modeling?}}

UnifiedAD separates transferable anomaly knowledge from target-specific normality modeling. Auxiliary normal and anomalous samples provide prior information about feature responses that are informative for anomaly discrimination, enabling the identification of anomaly neurons and anomaly directions. Such information captures discriminative properties that can be reused across categories and domains. In contrast, the normal subspace and calibration reference are constructed from a single normal observation of the target category, allowing the framework to characterize category-specific normal patterns without gradient-based adaptation. These two sources of information play complementary roles: auxiliary evidence guides the identification of informative feature responses, while the target normal reference determines how deviations from category-specific normality are evaluated. This separation provides a practical mechanism for transferring anomaly-related knowledge while retaining the flexibility of one-shot adaptation to previously unseen categories.

\noindent$\triangleright$ \textbf{\textit{Q3. Why Are Subspace and Auxiliary-Direction Evidence Complementary?}}

UnifiedAD characterizes anomalous patterns from two complementary perspectives. Subspace residuals measure deviations from the normal feature structures estimated using the target reference, capturing category-specific abnormalities without requiring target anomalous samples. In contrast, auxiliary-direction responses emphasize discriminative feature variations associated with anomalous patterns observed across source categories, introducing transferable anomaly-related information. These two forms of evidence therefore characterize different aspects of abnormality: one evaluates departures from target normality, while the other highlights responses associated with previously observed anomalies. The ablation results further reveal their complementary behavior, with auxiliary-direction evidence consistently improving image-level detection while exhibiting different effects on pixel-level localization across benchmarks. This suggests that combining reference-based deviations with transferable discriminative responses provides a more comprehensive assessment of anomalous patterns than relying on either perspective alone.

\noindent$\triangleright$ \textbf{\textit{Q4. How Does Calibrated Anomaly Evidence Enable Cross-Modal Fusion?}}

A central challenge in multi-modal anomaly detection is that feature deviations from different modalities may exhibit substantially different numerical scales and normal variation patterns. UnifiedAD addresses this challenge by calibrating modality-specific deviations against their corresponding normal reference residual distributions. Through empirical-tail calibration, raw feature deviations are transformed into reference-relative anomaly evidence, providing a common interpretation of how unusual each response is with respect to its modality-specific normal variations. Consequently, RGB and geometric features can be modeled independently while their calibrated anomaly evidence is aggregated within a shared inference framework. This enables appearance-related and geometry-related deviations to complement each other without requiring explicit alignment of their feature distributions. The improvements observed through RGB--geometry fusion further support the effectiveness of combining heterogeneous anomaly cues at the evidence level, highlighting normal-variation calibration as an important mechanism for unified multi-modal inference.

\noindent$\triangleright$ \textbf{\textit{Q5. Why Do Some Anomalies Remain Undetected?}}

The remaining failure cases cannot be solely attributed to insufficient coverage of normal variations under the one-shot setting. Extremely small, low-contrast, or subtle geometric defects may produce weak responses in frozen patch-level representations, causing anomalous features to overlap with normal variations. Although additional normal references can improve normality estimation, they cannot recover anomaly information lost during feature extraction. Moreover, a more comprehensive normal subspace may partially absorb anomalous variations that align with normal variation directions, resulting in small projection residuals. The transferability of auxiliary anomaly knowledge also introduces limitations, as neuron selections and anomaly directions learned from source domains may respond weakly to substantially different target anomalies. Finally, empirical-tail calibration may assign low anomaly evidence to deviations within the normal residual range, while fixed fusion weights may further suppress weak layer-specific or modality-specific cues. These factors jointly explain why increasing normal support samples alone cannot eliminate all missed detections.

\end{document}